\documentclass[prodmode,acmcsur]{acmsmall} % Aptara syntax

\usepackage[ruled]{algorithm2e}
\usepackage{amsmath}

\usepackage{array}
\newcolumntype{L}{>{\centering\arraybackslash}m{3cm}}
\usepackage{booktabs}
\usepackage{multirow}
\usepackage{makecell}
\usepackage{graphicx}
\usepackage{soul}
 \usepackage{color,soul}

\SetAlFnt{\small}
\SetAlCapFnt{\small}
\SetAlCapNameFnt{\small}
\SetAlCapHSkip{0pt}
\IncMargin{-\parindent}

\usepackage{subcaption}
\usepackage{float,lscape}

\acmVolume{X}
\acmNumber{X}
\acmArticle{1}
\acmYear{2017}
\acmMonth{8}

\setcopyright{acmcopyright}

\doi{0000001.0000001}

\issn{1234-56789}

\begin{document}

% Page heads
\markboth{P. Wang et al.}{Machine Learning for Survival Analysis: A Survey}

% Title portion
\title{Machine Learning for Survival Analysis: A Survey}
\author{
PING WANG
\affil{Virginia Tech}
YAN LI，
\affil{University of Michigan, Ann Arbor}
CHANDAN K. REDDY
\affil{Virginia Tech}
}
% NOTE! Affiliations placed here should be for the institution where the
%       BULK of the research was done. If the author has gone to a new
%       institution, before publication, the (above) affiliation should NOT be changed.
%       The authors 'current' address may be given in the "Author's addresses:" block (below).
%       So for example, Mr. Abdelzaher, the bulk of the research was done at UIUC, and he is
%       currently affiliated with NASA.

\begin{abstract}
{Survival analysis is a subfield of statistics where the goal is to analyze and model the data where the outcome is the time until the occurrence of an event of interest.} One of the main challenges in this context is the presence of instances whose event outcomes become unobservable after a certain time point or when some instances do not experience any event during the monitoring period. Such a phenomenon is called censoring which can be effectively handled using survival analysis techniques. Traditionally, statistical approaches have been widely developed in the literature to overcome this censoring issue. In addition, many machine learning algorithms are adapted to effectively handle survival data and tackle other challenging problems that arise in real-world data. In this survey, we provide a comprehensive and structured review of the representative statistical methods along with the machine learning techniques used in survival analysis and provide a detailed taxonomy of the existing methods. We also discuss several topics that are closely related to survival analysis and illustrate several successful applications in various real-world application domains. We hope that this paper will provide a more thorough understanding of the recent advances in survival analysis and offer some guidelines on applying these approaches to solve new problems that arise in applications with censored data.	

\end{abstract}

%
% The code below should be generated by the tool at
% http://dl.acm.org/ccs.cfm
% Please copy and paste the code instead of the example below.
%


%[ADD ONE MORE BUT NOT ABOUT HEALTHCARE]
%\ccsdesc[500]{Applied computing~Health care information systems}
%
% End generated code
%

%\terms{Design, Algorithms, Performance}


%\acmformat{Ping Wang, Yan Li, Chandan K. Reddy, 2017. Machine Learning for Survival Analysis: A Survey.}
% At a minimum you need to supply the author names, year and a title.
% IMPORTANT:
% Full first names whenever they are known, surname last, followed by a period.
% In the case of two authors, 'and' is placed between them.
% In the case of three or more authors, the serial comma is used, that is, all author names
% except the last one but including the penultimate author's name are followed by a comma,
% and then 'and' is placed before the final author's name.
% If only first and middle initials are known, then each initial
% is followed by a period and they are separated by a space.
% The remaining information (journal title, volume, article number, date, etc.) is 'auto-generated'.

\begin{bottomstuff}
This material is based upon work supported by, or in part by, the U.S. National Science Foundation grants IIS-1707498, IIS-1619028 and IIS-1646881.

Author's addresses: P. Wang is with the Department of Computer Science, Virginia Tech, 900 N. Glebe Road, Arlington, VA, 22203. E-mail: ping@vt.edu; Y. Li is with the Department of Computational Medicine and Bioinformatics, University of Michigan, 100 Washtenaw Avenue, Ann Arbor, MI, 48109. E-mail: yanliwl@umich.edu; C. K. Reddy is with the Department of Computer Science, Virginia Tech, 900 N. Glebe Road, Arlington, VA, 22203. E-mail: reddy@cs.vt.edu.
\end{bottomstuff}

\maketitle
\section{Introduction}
\label{sec:intro}
Due to the development of various data acquisition and big data technologies, the ability to collect a wide variety of data and monitor the observation over long-term periods have been attained in different disciplines. For most of the real-world applications, the primary objective of monitoring these observations is to obtain a better estimate of the time of occurrence of a particular event of interest. One of the main challenges for such time-to-event data is that usually there exist censored instances, i.e., the event of interests is not observed for these instances due to either the time limitation of the study period or losing track during the {observation period}. 
%the data is often incomplete due to the fact that some instances in the longitudinal data will either not experience any event or become unobservable during the observation period. 
More precisely, certain instances have experienced event (or labeled as event) and the information about the outcome variable for the remaining instances is only available until a specific time point in the study. Therefore, it is not suitable to directly apply predictive algorithms using the standard statistical and machine learning approaches to analyze the {survival} data. Survival analysis, which is an important subfield of statistics, provides various mechanisms to handle such censored data problems that arise in modeling such {complex} data (also referred to as time-to-event data when modeling a particular event of interest is the main objective of the problem) which occurs ubiquitously in various real-world application domains.

In addition to the difficulty in handling the censored data, there are also several unique challenges to perform the predictive modeling with such survival data and hence several researchers have, more recently, developed new computational algorithms for effectively handling such complex challenges. To tackle such practical concerns, some related works have adapted several machine learning methods to solve the survival analysis problems and machine learning researchers have developed more sophisticated and effective algorithms which either complement or compete with the traditional statistical methods. In spite of the importance of these problems and relevance to various real-world applications, this research topic is scattered across different disciplines. Moreover, there are only few surveys that are available in the literature on this topic and, to the best of our knowledge, there is no comprehensive review paper about survival analysis and its recent developments from a machine learning perspective. Almost all of these existing survey articles describe solely statistical methods and either completely ignore or barely mention the machine learning advancements in this research field. One of the earliest surveys may be found in \cite{chung1991survival}, which gives an overview of the statistical survival analysis methods and describes its applications in criminology by predicting the time until recidivism. Most of the existing books about survival analysis~\cite{kleinbaum2006survival,lee2003statistical,allison2010survival} focus on introducing this topic from the traditional statistical perspective instead of explaining from the machine learning standpoint. Recently, the authors in~\cite{cruz2006applications} and~\cite{kourou2015machine} discussed the applications in cancer prediction and provided a comparison of several machine learning techniques.

The primary purpose of this survey article is to provide a comprehensive and structured overview of various machine learning methods for survival analysis along with the traditional statistical methods. We demonstrate the commonly used evaluation metrics and advanced related formulations that are commonly investigated in this research topic. We will discuss a detailed taxonomy of all the survival analysis methods that were developed in the traditional statistics as well as more recently in the machine learning community. We will also provide links to various implementations and sources codes which will enable the readers to further dwell into the methods discussed in this article. Finally, we will discuss various applications of survival analysis.

The rest of this paper is organized as follows. We will give a brief review of the basic concepts, notations and definitions that are necessary to comprehend the survival analysis algorithms and provide the formal problem statement for survival analysis problem in Section~\ref{sec2}. A taxonomy of the existing survival analysis methods, including both statistical and machine learning methods will also be provided to elucidate the holistic view of the existing works in the area of survival analysis. We will then review the well-studied representative conventional statistical methods including non-parametric, semi-parametric, and parametric models in Section~\ref{sec3}. Section~\ref{sec4} describes several basic machine learning approaches, including survival trees, Bayesian methods, support vector machines and neural networks developed for survival analysis. Different kinds of advanced machine learning algorithms such as ensemble learning, transfer learning, multi-task learning and active learning for handling survival data will also be discussed. Section~\ref{sec5} demonstrates the evaluation metrics for survival models. In addition to the survival analysis algorithms, some interesting topics related to this topic have received considerable attention in various fields. In Section~\ref{sec6}, several related concepts such as early prediction and complex events will be discussed. Various data transformation techniques such as uncensoring and calibration which are typically used in conjunction with existing predictive methods will also be mentioned briefly. A discussion about topics in complex event analysis such as competing risks and recurrent events will also be provided. In Section~\ref{sec7}, various real-world applications of survival analysis methods will be briefly explained and more insights into these application domains will be provided. In Section~\ref{sec8}, the details about the implementations and software packages of the survival analysis methods are discussed. Finally, Section~\ref{sec9} concludes our discussion.% and \textbf{provides brief discussion about ongoing efforts and potential future topics related to this topic}.

\section{Definition of Survival Analysis}
\label{sec2}
In this section, we will first provide the basic notations and terminologies used in this paper. We will then give an illustrative example which explains the structure of the survival data and give a more formal problem statement for survival analysis. At last, we also give a complete taxonomy of the existing survival analysis methods that are available in the literature, including both the conventional statistical methods and the machine learning approaches. It provides a holistic view of the field of survival analysis and will aid the readers to gain the basic knowledge about the methods used in this field before getting into the detailed algorithms.
%\vspace{-2mm}
\begin{table}[h]
	\tbl{Notations used in this paper.\label{tab:notations}}{%
		\def\arraystretch{1.3}
		\begin{tabular}{|c|l|}
			\hline
			\textbf{Notations}  & \textbf{Descriptions} \\\hline	
			\hline
			$P$ &\multicolumn{1}{m{5cm}|}{The number of features} \\\hline
			$N$ &\multicolumn{1}{m{5cm}|}{The number of instances} \\\hline
			
			$X$ & $ \mathbb{R}^{N \times P}$ feature vector \\\hline
			$X_i$ & $\mathbb{R}^{1\times P}$ covariate vector of instance $i$ \\\hline
			$T$ & $\mathbb{R}^{N\times 1}$ vector of event times \\\hline
			$C$ & $\mathbb{R}^{N\times 1}$ vector of last follow up times \\\hline
			$y$ & $\mathbb{R}^{N\times 1}$ vector of observed time which is equal to $min(T,C)$ \\\hline
			$\delta$ & $N\times 1$ binary vector for event status \\\hline
			$\beta$ & $\mathbb{R}^{P\times 1}$ coefficient vector \\\hline
			$f(t)$ & Death density function \\\hline
			$F(t)$ & Cumulative event probability function \\\hline
			$S(t)$ & Survival probability function \\\hline	
			$h(t)$ & Hazard function \\\hline
			$h_0(t)$ & Baseline hazard function \\\hline	
			$H(t)$ & Cumulative hazard function \\\hline		
		\end{tabular}}
\end{table}%
%\vspace{-4mm}
\subsection{Survival Data and Censoring}
\label{sec:1.1}
During the study of a survival analysis problem, it is possible that the events of interest are not observed for some instances; this scenario occurs because of the limited observation time window or missing traces caused by other uninterested events. This concept is known as censoring \cite{klein2005survival}. {We can broadly categorize censoring into three groups based on the occurrence of the censoring}~\cite{lee2003statistical}{, (i) \textit{right-censoring}, for which the observed survival time is less than or equal to the true survival time; (ii) \textit{left-censoring}, for which the observed survival time is greater than or equal to the true survival time; and (iii) \textit{interval censoring},  for which we only know that the event occurs during a given time interval. It should be noted that the true event occurrence time is unknown in all the three cases. Among them, right-censoring is the most common scenario that arises in many practical problems}~\cite{marubini2004analysing}, {thus, the survival data with right-censoring information will be mainly analyzed in this paper.}

For a survival problem, the time to the event of interest $(T)$ is known precisely only for those instances who have the event occurred during the study period. For the remaining instances, since we may lose track of them during the observation time or their time to event is greater than the observation time, we can only have the censored time $(C)$ which may be the time of withdrawn, lost or the end of the observation. They are considered to be censored instances in the context of survival analysis. In other words, here, we can only observe either survival time $(T_i)$ or censored time $(C_i)$ but not both, for any given instance $i$. If and only if $y_i=min(T_i,C_i)$ can be observed during the study, the dataset is said to be right-censored{. In the survival problem with right-censored instances, the censoring time is also a random variable since the instances enter the study randomly and the randomness in the censoring time of the instances. Thus, in this paper, we assume that the censoring occurs randomly in the survival problems. For the sake of brevity, we will refer to the randomly occurred right-censoring as \textit{censoring} hereafter in the paper.}%, which is a common scenario that arises in many practical problems~\cite{marubini2004analysing}.

\begin{figure*}[h]
	\centering
	%\scalebox{.5}{\includegraphics{example}}
	\scalebox{.4}{\includegraphics[trim={0cm 0cm 0cm 4.5cm},clip]{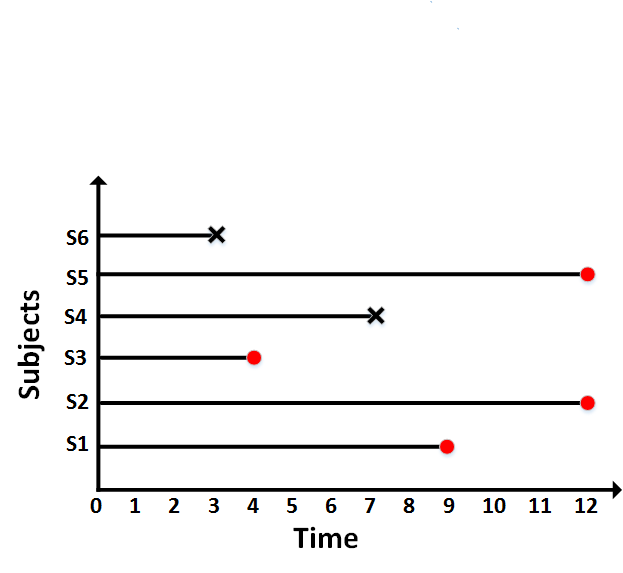}} \hspace{-0.1 cm}
	\caption{An illustration demonstrating the survival analysis problem.}
	\label{Fig:data}
\end{figure*}

In Figure~\ref{Fig:data}, an illustrative example is given for a better understanding of the definition of censoring and the structure of survival data. Six instances are observed in this {study} for $12$ months and the event occurrence information during this time period is recorded. From Figure~\ref{Fig:data}, we can find that only subjects S4 and S6 have experienced the event (marked by `X') during the follow-up time and the observed time for them is the event time. While the event did not occur within the $12$ months period for subjects S1, S2, S3 and S5, which are considered to be censored and marked by red dots in the figure. More specifically, subjects S2 and S5 are censored since there was no event occurred during the study period, while subjects S1 and S3 are censored due to the withdrawal or being lost to follow-up within the study time period.
%For a better understanding of the structure of survival data and the definition of censoring, let us consider an illustrative example shown in Figure \ref{Fig:data}. In this example, a longitudinal study for $12$ months is conducted on six subjects and the information for event occurrence until the ending time is recorded, where only S4 and S6 have experienced the event (marked by `X') and the observed time for them is the event time. While for S1, S2, S3 and S5 the event did not occurred within the $12$ months period are considered to be censored at the end of the study and marked by red dots. More specifically, subjects S2 and S5 are censored since they did not experience the event before the end of the study period, while subjects S1 and S3 are censored during the follow-up time period due to the withdrawal from the study or being lost to follow-up.
\vspace{-1.8mm}

%In other words, we can only observe the censored time for the censored subjects, for example, $y_3=4$ for subject S3 and $y_5=12$ for subject S5.
%\vspace{-4mm}
%In survival analysis, survival data are normally represented by a triple of variables $(X,Z,\delta)$, where $X$ is the feature vector, and $\delta$ is an indicator. $\delta=1$ if $Z$ is the time to the event of interest and $\delta=0$ if $Z$ is the censored time; for convenience, $Z$ is usually named the \emph{observed time} \cite{lee2003statistical}. 

\textbf{Problem Statement:}
For a given instance $i$, represented by a triplet $(X_i,y_i,\delta_i)$, where $X_i\in\mathbb{R}^{1\times P}$ is the feature vector; $\delta_i$ is the binary event indicator, i.e., $\delta_i=1$ for an uncensored instance and $\delta_i=0$ for a censored instance; and $y_i$ denotes the observed time and is equal to the survival time $T_i$ for an uncensored instance and $C_i$ for a censored instance, i.e., 
\begin{equation}
y_i=\begin{cases}
T_i & \text{if}\ \delta_i=1\\
C_i & \text{if}\ \delta_i=0
\end{cases}
\end{equation}
It should be noted that $T_i$ is a latent value for censored instances since these instances did not experience any event during the observation time period.
%As discussed above, for the $i^{th}$ instance in the survival data, the time-to-event of interest and the censored time can be denoted as $T_i$ and $C_i$, respectively. It should be noted that $C_i$ is a latent value for censored instances since the event does not occur during the observation time period. In this case, the observed time $y_i$ for the $i^{th}$ instance is defined as follows:
%\begin{equation}
%y_i=\begin{cases}
%T_i & \text{if event occurs for instance}\ i.\\
%C_i & \text{if instance}\ i\ \text{is censored}.
%\end{cases}
%\end{equation}
%In survival analysis, each instance can be represented by a triplet $(X_i,y_i,\delta_i)$, where $X_i\in\mathbb{R}^{1\times P}$ is the feature vector for the $i^{th}$ instance, and $\delta_i$ is the binary event indicator. $\delta_i=1$ if $y_i=T_i$, and $\delta_i=0$ if $y_i=C_i$.

The goal of survival analysis is to estimate the time to the event of interest $T_j$ for a new instance $j$ with feature predictors denoted by $X_j$. It should be noted that, in survival analysis problem, the value of $T_j$ will be both non-negative and continuous.

\subsection{Survival and Hazard Function}
The \textbf{\emph{survival function}}, which is used to represent the probability that the time to the event of interest is not earlier than a specified time $t$ \cite{lee2003statistical,klein2005survival}, is one of the primary goals in survival analysis. Conventionally, survival function is represented by $S$, which is given as follows:
\begin{equation}
\label{equ:SF}
S(t)=Pr(T\ge t).
\end{equation}
The survival function monotonically decreases with $t$, and the initial value is 1 when $t=0$, which represents the fact that, in the beginning of the observation, $100\%$ of the observed subjects survive; in other words, none of the events of interest have occurred.

\begin{figure*}[h]
	\centering
	\scalebox{.4}{\includegraphics[trim={0cm 0cm 0cm 1.1cm},clip]{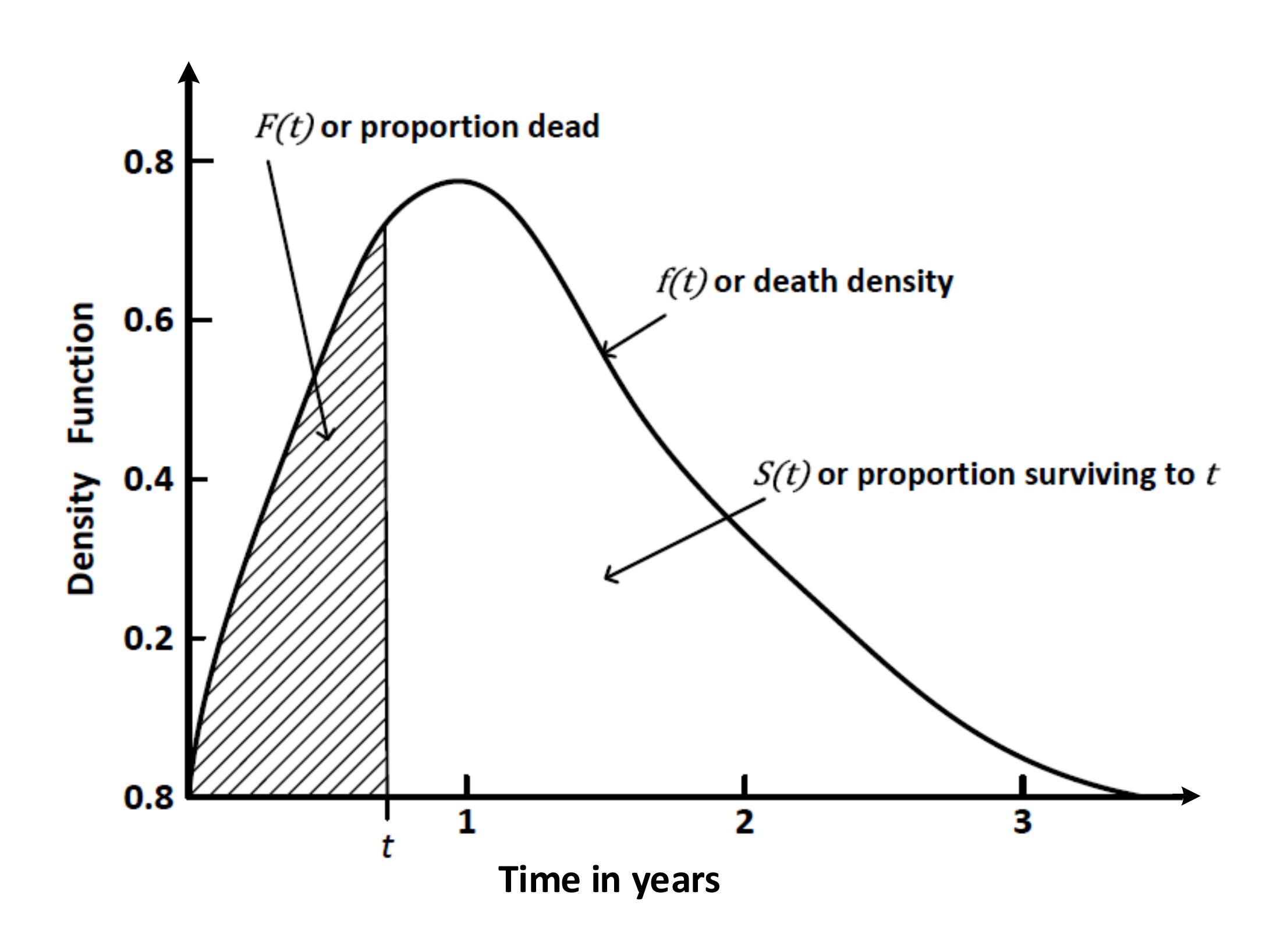}}
	\caption{Relationship among different entities $f(t)$, $F(t)$ and $S(t)$.}
	\label{Fig:fsf}
	\vspace{-6.2mm}
\end{figure*}

On the contrary, the \emph{cumulative death distribution function} $F(t)$, which represents the probability that the event of interest occurs earlier than $t$, is defined as $F(t)=1-S(t)$, and \textbf{\emph{death density function}} can be obtained as $f(t)=\frac{d}{dt}F(t)$ for continuous cases, and $f(t)=[{F(t+\Delta t)-F(t)}]/{\Delta t}$, where $\Delta t$ denotes a small time interval, for discrete cases. Figure~\ref{Fig:fsf} shows the relationship among these functions.
\vspace{-3.2mm}
%\vspace{-4mm}

In survival analysis, another commonly used function is the \textbf{\emph{hazard function}} ($h(t)$), which is also called the \emph{force of mortality}, the \emph{instantaneous death rate} or the \emph{conditional failure rate} \cite{dunn2009basic}. The hazard function does not indicate the chance or probability of the event of interest, but instead it is the rate of event at time $t$ given that no event occurred before time $t$. Mathematically, the hazard function is defined as:
\begin{eqnarray}
\label{eqn:HF}
h(t)&=&\lim_{\Delta t\to 0}\frac{Pr(t\le T <t+\Delta t~|~T\ge t)}{\Delta t}=\lim_{\Delta t\to 0}\frac{F(t+\Delta t)-F(t)}{\Delta t\cdot S(t)}=\frac{f(t)}{S(t)}
\end{eqnarray}
Similar to $S(t)$, $h(t)$ is also a non-negative function. While all the survival functions, $S(t)$, decrease over time, the hazard function can have a variety of shapes. Consider the definition of $f(t)$, which can also be expressed as $f(t)=-\frac{d}{dt}S(t)$, so the hazard function can be represented as:
\begin{equation}
h(t)=\frac{f(t)}{S(t)}=-\frac{d}{dt}S(t)\cdot\frac{1}{S(t)}=-\frac{d}{dt}[lnS(t)].
\end{equation}
Thus, the survival function defined in Eq.~(\ref{equ:SF}) can be rewritten as
\begin{equation}
S(t)=exp(-H(t))
\end{equation}
where $H(t)=\int_0^t h(u)du$ represents the \emph{cumulative hazard function} (CHF)~\cite{lee2003statistical}.

\subsection{Taxonomy of Survival Analysis methods}
Broadly speaking, the survival analysis methods can be classified into two main categories: statistical methods and machine learning based methods. Statistical methods share the common goal with machine learning methods to {make predictions of the survival time and estimate the survival probability at the estimated survival time}. However, they focus more on {characterizing} both the distributions of the event times and the statistical properties of the parameter estimation{ by estimating the survival curves, while machine learning methods focus more on the prediction of event occurrence at a given time point by incorporating the traditional survival analysis methods with various machine learning techniques}. 
%The main advantages of machine learning methods are that there is no underlying assumption for neither the distribution of the event times nor the relationships between covariates and event times. In addition, 
Machine learning methods are usually applied to the high-dimensional problems, while statistical methods are generally developed for the low-dimensional data. In addition, machine learning methods for survival analysis offer more effective algorithms by incorporating survival problems with both statistical methods and machine learning methods and taking advantages of the recent developments in machine learning and optimization to learn the dependencies between covariates and survival times in different ways. 

\begin{figure*}[!htp]
	%\centering
	\hspace*{-1in}
	\includegraphics[width=7.1in]{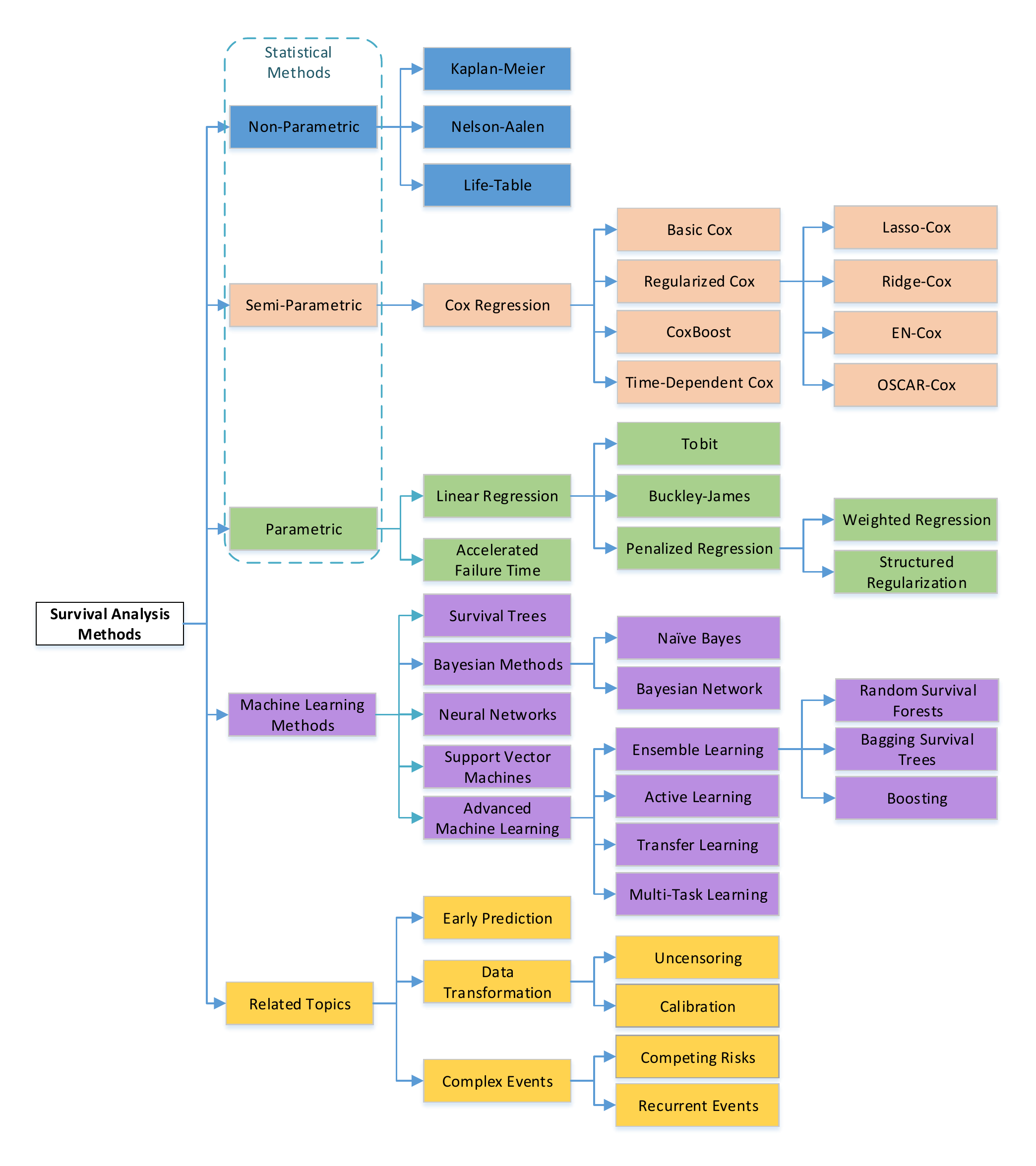}
	\vspace{-8mm}
	\caption{Taxonomy of the methods developed for survival analysis.}
	\label{Fig:taxonomy}
	%\vspace{-5mm}
\end{figure*}

Based on the assumptions and the usage of the parameters used in the model, the traditional statistical methods can be subdivided into three categories: (i) non-parametric models, (ii) semi-parametric models and (iii) parametric models. Machine learning algorithms, such as survival trees, Bayesian methods, neural networks and support vector machines, which have become more popular in the recent years are included under a separate branch. Several advanced machine learning methods, including ensemble learning, active learning, transfer learning and multi-task learning methods, are also included. The overall taxonomy also includes some of the research topics that are related to survival analysis such as complex events, data transformation and early prediction. A complete taxonomy of these survival analysis methods is shown in Figure \ref{Fig:taxonomy}.

%\vspace{-2mm}
\section{Traditional statistical methods} 
\label{sec3}
In this section, we will introduce three different types of statistical methods to estimate the survival/hazard functions: non-parametric, semi-parametric and parametric methods. Table \ref{tab:one} shows both the advantages and disadvantages of each type of methods {based on theoretical and experimental analysis} and lists the specific methods in each type.
%\vspace{-2mm}
\begin{table}[b]%
\tbl{Summary of different types of statistical methods for survival analysis.\label{tab:one}}{%
\begin{tabular}{|c|c|c|c|}
			\hline
			\textbf{Type}  & \textbf{Advantages} &\textbf{Disadvantages} &\textbf{Specific methods}\\\hline	
			\hline
			Non-parametric&\multicolumn{1}{m{3.2cm}|}{More efficient when no suitable theoretical distributions known.} &\multicolumn{1}{m{3.2cm}|}{Difficult to interpret; yields inaccurate estimates.}  &\thead{Kaplan-Meier\\ Nelson-Aalen \\Life-Table}\\\hline
			
			Semi-parametric &\multicolumn{1}{m{3.2cm}|}{The knowledge of the underlying distribution of survival times is not required.}   &\multicolumn{1}{m{3.2cm}|}{The distribution of the outcome is unknown; not easy to interpret.} &\thead{Cox model\\ Regularized Cox\\ CoxBoost\\ Time-Dependent Cox}\\\hline
			
			Parametric &\multicolumn{1}{m{3.2cm}|}{Easy to interpret, more efficient and accurate when the survival times follow a particular distribution.}  &\multicolumn{1}{m{3.2cm}|}{When the distribution assumption is violated, it may be inconsistent and can give sub-optimal results.}  &\thead{Tobit\\ Buckley-James\\ Penalized regression\\ Accelerated Failure Time} \\\hline			
\end{tabular}}
\end{table}%
%\vspace{-2mm}

{Non-parametric methods are more efficient when there is no underlying distribution for the event time or the proportional hazard assumption does not hold.} In non-parametric methods, an empirical estimate of the survival function is obtained using Kaplan-Meier (KM) method, Nelson-Aalen estimator (NA) or Life-Table (LT) method. More generally, any KM estimator for the survival probability at the specified survival time is a product of the same estimate up to the previous time and the observed survival rate for that given time. Thus, KM method is also referred to as a \emph{product-limit method} \cite{kaplan1958nonparametric,lee2003statistical}. NA method is an estimator based on modern counting process techniques~\cite{andersen2012statistical}. LT \cite{cutler1958maximum} is the application of the KM method to the interval grouped survival data.

Under the semi-parametric category, Cox model is the most commonly used regression analysis approach for survival data and {it differs significantly from other methods since} it is built on the proportional hazards assumption and employs partial likelihood for parameter estimation. Cox regression method is described as semi-parametric method since the distribution of the outcome remains unknown even if it is based on a parametric regression model. In addition, several useful variants of the basic Cox model, such as penalized Cox models, CoxBoost algorithm and Time-Dependent Cox model (TD-Cox), are also proposed in the literature.

Parametric methods are more efficient and accurate for estimation when the time to event of interest follows a particular distribution specified in terms of certain parameters. It is relatively easy to estimate the times to the event of interest with parametric models, but  it becomes awkward or even impossible to do so with the Cox model~\cite{allison2010survival}. Linear regression method is one of the main parametric survival methods, while the Tobit model, Buckley-James regression model and the penalized regression are the most commonly used linear models for survival analysis. In addition, other parametric models, such as accelerated failure time (AFT) which models the survival time as a function of covariates \cite{kleinbaum2006survival}, are also widely used. We will now describe these three types of statistical survival methods in this section.
%\vspace{-3.5mm}

\subsection{Non-parametric Models}
%\textbf{If no suitable theoretical distribution is known, nonparametric methods are more efficient. (move this later)}

Among all functions, the survival function or its graphical presentation %the survival curve 
is the most widely used one. In 1958, Kaplan and Meier \cite{kaplan1958nonparametric} developed the Kaplan-Meier (KM) Curve or the product-limit (PL) estimator to estimate the survival function using the actual length of the observed time. This method is the most widely used one for estimating survival function. Let $T_1<T_2<\cdots<T_K$ be a set of distinct ordered event times observed for $N (K\leq N)$ instances. In addition to these event times, there are also censoring times for instances whose event times are not observed. For a specific event time $T_j~(j=1,2,\cdot \cdot \cdot, K)$, the number of observed events is $d_j\geq 1$, and $r_j$ instances will be considered to be ``at risk" since their event time or censored time is greater than or equal to $T_j$. It should be noted that we cannot simply consider $r_j$ as the difference between $r_{j-1}$ and $d_{j-1}$ due to the censoring. The correct way to obtain $r_j$ is $r_j=r_{j-1}-d_{j-1}-c_{j-1}$, where $c_{j-1}$ is the number of censored instances during the time period between $T_{j-1}$ and $T_{j}$. Then the conditional probability of surviving beyond time $T_j$ can be defined as:
\begin{equation}
\label{Equ:cpsT}
p(T_j)=\frac{r_j-d_j}{r_j}
\end{equation}
Based on this conditional probability, the product-limit estimate of survival function $S(t)=P(T \ge t)$ is given as follows:
\begin{equation}
\hat{S}(t)=\prod_{j:T_j<t}p(T_j)=\prod_{j:T_j<t}(1-\frac{d_j}{r_j})
\end{equation}

However, if the subjects in the data are grouped into some interval periods according to the time, or if the number of subjects is very large, or when the study is for a large population, the Life Table (LT) analysis~\cite{cutler1958maximum} will be a more convenient method. Different from KM and LT method, Nelson-Aalen estimator~\cite{nelson2000theory,aalen1978nonparametric} is a method to estimate the cumulative hazard function for censored data based on counting process approach. It should be noted that when the time to event of interest follows a specific distribution, nonparametric methods are less efficient compared to the parametric methods.

\subsection{Semi-Parametric Models}
As a hybrid of the parametric and non-parametric approaches, semi-parametric models can obtain a more consistent estimator under a broader range of conditions compared to the parametric models, and a more precise  estimator than the non-parametric methods~\cite{powell1994estimation}. Cox model~\cite{cox1972regression} is the most commonly used survival analysis method in this category. Unlike parametric methods, the knowledge of the underlying distribution of time to event of interest is not required, but the attributes are assumed to have an exponential influence on the outcome. We will now discuss the details of Cox model more elaborately and then describe different variants and extensions of the basic Cox model such as regularized Cox models, CoxBoost and Time-Dependent Cox.

\subsubsection{The Basic Cox Model}
\label{sec:cox} For a given instance $i$, represented by a triplet $(X_i,y_i,\delta_i)$, the hazard function $h(t,X_i)$ in the Cox model follows the proportional hazards assumption given by 
\begin{equation}
\label{equ:cox}
h(t,X_i)=h_0(t)exp(X_i\beta),
\end{equation}
for $i=1,2,\cdots,N$, where the \emph{baseline hazard function}, $h_0(t)$, can be an arbitrary non-negative function of time, $X_i=(x_{i1},x_{i2},\cdots,x_{iP})$ is the corresponding covariate vector for instance $i$, and $\beta^T=(\beta_1,\beta_2,\cdots,\beta_P)$ is the coefficient vector. The Cox model is a semi-parametric algorithm since the baseline hazard function, $h_0(t)$, is unspecified. For any two instances $X_1$ and $X_2$, the hazard ratio is given by
\begin{equation}
	\frac{h(t,X_1)}{h(t,X_2)}=\frac{h_0(t)exp(X_1\beta)}{h_0(t)exp(X_2\beta)}=exp[(X_1-X_2)\beta].
\end{equation}
which means that the hazard ratio is independent of the baseline hazard function. Cox model is a proportional hazards model since the hazard ratio is a constant and all the subjects share the same baseline hazard function. Based on this assumption, the survival function can be computed as follows:
\begin{equation}
S(t)=exp(-H_0(t)exp(X\beta))=S_0(t)^{exp(X\beta)}
\end{equation}
where $H_0(t)$ is the \emph{cumulative baseline hazard function}, and $S_0(t)=exp(-H_0(t))$ represents the baseline survival function.
%However, in order to have the survival function, we need to estimate $\Lambda_0(t)$ and $S_0(t)=exp(-\Lambda_0(t))$ first. 
The Breslow's estimator~\cite{breslow1972} is the most widely used method to estimate $H_0(t)$, which is given by 
\begin{equation}
\hat{H}_0(t)=\sum_{t_i \le t}\hat{h}_0(t_i) 
\end{equation}
where $\hat{h}_0(t_i)=1/\sum_{j \in R_i}e^{X_j\beta }$ if $t_i$ is an event time, otherwise $\hat{h}_0(t_i)=0$. Here, $R_i$ represents the set of subjects who are at risk at time $t_i$.
%\begin{equation}
%\hat{\lambda}_0(t_i)= \begin{cases}
%1/\sum_{j \in R_i}e^{\beta x_j} & if\ is\ an\ event\ time\\
%0 & otherwise
%\end{cases}
%\end{equation}
%\subsubsection{Estimation of the Regression Parameters}\

Because the baseline hazard function $h_0(t)$ in Cox model is not specified, it is not possible to fit the model using the standard likelihood function. In other words, the hazard function $h_0(t)$ is a nuisance function, while the coefficients $\beta$ are the parameters of interest in the model. To estimate the coefficients, Cox proposed a partial likelihood~\cite{cox1972regression,cox1975partial} which depends only on the parameter of interest $\beta$ and is free of the nuisance parameters. The hazard function refers to the probability that an instance with covariate $X$ fails at time $t$ on the condition that it survives until time $t$ can be expressed by $h(t,X)dt$ with $dt\to 0$. Let $J\ (J\leq N)$ be the total number of events of interest that occurred during the observation period for $N$ instances, and $T_1<T_2<\cdots<T_J$ is the distinct ordered time to event of interest. Without considering the ties, let $X_j$ be the corresponding covariate vector for the subject who fails at $T_j$, and $R_j$ be the set of risk subjects at $T_j$. Thus, conditional on the fact that the event occurs at $T_j$, the  individual probability corresponding to covariate $X_j$ can be formulated as follows:
%\vspace{-2mm}
\begin{equation}
\frac{h(T_j,X_j)dt}{\sum_{i\in R_j}h(T_j,X_i)dt}
\end{equation}
and the partial likelihood is the product of the probability of each subject; referring to the Cox assumption and the presence of the censoring, the partial likelihood is defined as follows:
%\vspace{-2mm}
\begin{equation}
L(\beta)=\prod_{j=1}^N\left[\frac{exp(X_j\beta)}{\sum_{i\in R_j}exp(X_i\beta)}\right]^{\delta_j}
\end{equation}
It should be noted that here $j=1,2,\cdots,N$; if $\delta_j=1$, the $j^{th}$ term in the product is the conditional probability; otherwise, when $\delta_j=0$, the corresponding term is $1$, which means that the term will not have any effect on the final product. The coefficient vector $\hat{\beta}$ is estimated by maximizing this partial likelihood, or equivalently, minimizing the negative \emph{log-partial likelihood} for improving efficiency.
%\vspace{-2mm}
\begin{equation}
\label{equ:lpl}
LL(\beta)=-\sum_{j=1}^N\delta_j\{X_j\beta-log[\sum_{i\in R_j}exp(X_i\beta)]\}.
\end{equation}
The maximum partial likelihood estimator (MPLE)~\cite{cox1972regression,lee2003statistical} can be used along with the numerical Newton-Raphson method~\cite{kelley1999iterative} to iteratively find an estimator $\hat{\beta}$ which minimizes $LL(\beta)$ {with time complexity $O(NP^2)$}.

\subsubsection{Regularized Cox models} 
With the development of data collection and detection techniques, most real-world domains tend to encounter {high-dimensional data}. In some cases, the number of variables ($P$) in the given data is almost equal to or even exceeds the number of instances ($N$). It is challenging to build the prediction model with all the features and the model might provide inaccurate results because of the overfitting problem~\cite{van2011dynamic}. This motivates using sparsity norms to select vital features in high-dimension under the assumption that most of the features are not significant~\cite{friedman2001elements}. For the purpose of identifying the most relevant features to the outcome variable among tens of thousands of features, different penalty functions, including lasso, group lasso, fused lasso and graph lasso, are also used to develop the prediction models using the sparse learning methods. The family of $\ell$-norm penalty functions $\ell_{\gamma}: \mathbb{R}^{\gamma} \to \mathbb{R}$, with the form of  
%\begin{equation}
%\label{Equ:lp}
$\ell_{\gamma}(\beta)=\parallel \beta \parallel_{\gamma}=(\sum_{i=1}^P{\|\beta_i\|^{\gamma}})^{\frac{1}{\gamma}}, \gamma>0$
%\end{equation}
are the commonly used penalty functions. The smaller the value of $\gamma$, the sparser the solution, but when $0\leq \gamma<1$, the penalty is non-convex, which makes the optimization problem more challenging to solve. Here, we will introduce the commonly used regularized Cox models, whose regularizers are summarized in Table \ref{tab:regularizers}.

\begin{table}[h]
\setlength\extrarowheight{5pt}
\centering	
\tbl{Different regularizers used in the variants of Cox model.\label{tab:regularizers}}{%
	\def\arraystretch{1.1}
\begin{tabular}{|c|c|c|c|}
\hline
\textbf{Regularized Cox models} & \textbf{Regularizers} \\\hline	
\hline
Lasso-Cox&$\lambda\sum_{p=1}^P|\beta_p|$ \\\hline
Ridge-Cox&$\frac{\lambda}{2}\sum_{p=1}^P\beta_p^2$\\\hline
EN-Cox&$\lambda[\alpha\sum_{p=1}^P|\beta_p|+\frac{1}{2}(1-\alpha)\sum_{p=1}^P\beta_p^2]$\\\hline
%KEN-Cox&$\lambda\alpha\parallel\beta\parallel_1+\lambda(1-\alpha)\beta^TK\beta,with\ K_{ij}=exp(\frac{-\parallel x_i-x_j\parallel_2^2}{2\sigma^2})$\\\hline
OSCAR-Cox&$\lambda_1\parallel\beta\parallel_1+\lambda_2\parallel T\beta\parallel_1$\\\hline
%GLasso-Cox&$\lambda \sum_{k=1}^{K}\sqrt{\sum_{j=1}^{p_k}\beta_{kj}^2}$\\\hline						
\end{tabular}}	
\vspace{-1mm}		
\end{table}%

\vspace{2mm}
\textbf{Lasso-Cox:} Lasso~\cite{tibshirani1996regression} is a $\ell_1$-norm regularizer which is good at performing feature selection and estimating the regression coefficients simultaneously. In~\cite{tibshirani1997lasso}, the $\ell_1$-norm penalty was incorporated into the log-partial likelihood shown in Eq. ~(\ref{equ:lpl}) to obtain the Lasso-Cox algorithm, which inherits the properties of $\ell_1$-norm in feature selection.

There are also some extensions of Lasso-Cox method. Adaptive Lasso-Cox~\cite{zhang2007adaptive} is based on a penalized partial likelihood with adaptively weighted $\ell_1$ penalties $\lambda\sum_{j=1}^P\tau_j|\beta_j|$ on regression coefficients, with small weights $\tau_j$ for large coefficients and large weights for small coefficients. In fused Lasso-Cox~\cite{tibshirani2005sparsity}, the coefficients and their successive differences are penalized using the $\ell_1$-norm. In graphical Lasso-Cox~\cite{friedman2008sparse}, the sparse graphs are estimated using coordinate descent method by applying a $\ell_1$-penalty to the inverse covariance matrix. These extensions solve the survival problems in a similar way as the regular Lasso-Cox model by incorporating different $\ell_1$ penalties.
\vspace{2mm}
%\begin{equation}
%\label{coxlasso}
%\hat{\beta}_{lasso}=\min_\beta\{-\frac{2}{N}[\sum_{j=1}^N\delta_j X_j\beta-\delta_j log(\sum_{i\in R_j}e^{X_i\beta})]+\lambda\sum_{p=1}^P|\beta_p|\}
%\end{equation}

\textbf{Ridge-Cox:} Ridge regression was originally proposed by Hoerl and Kennard~\cite{hoerl1970ridge} and was successfully used in the context of Cox regression by Verweij et al. \cite{verweij1994penalized}. It incorporates a $\ell_2$-norm regularizer to select the correlated features and shrink their values towards each other. 

Feature-based regularized Cox method (FEAR-Cox)~\cite{vinzamuri2013cox} uses feature-based non-negative valued regularizer $R(\beta)=|\beta|^TM|\beta|$ for the modified least squares formulation of Cox regression and the cyclic coordinate descent method is used to solve this optimization problem, where $M\in \mathbb{R}^{P \times P}$ $(P\ \text{is the number of features})$ is a positive semi-definite matrix. Ridge-Cox is a special case of FEAR-Cox when $M$ is the identity matrix. %$J(\beta)=\parallel\beta\parallel^2$.
\vspace{2mm}

%The regression parameters of Ridge-Cox can be estimated by
%\begin{equation}
%\label{coxridge}
%\hat{\beta}_{ridge}=\min_\beta\{-\frac{2}{N}[\sum_{j=1}^N\delta_j X_j\beta-\delta_j log(\sum_{i\in R_j}e^{X_i\beta})]+\frac{\lambda}{2}\sum_{p=1}^P\beta_p^2\}
%\end{equation}

\textbf{EN-Cox:} Elastic net (EN), which combines the $\ell_1$ and squared $\ell_2$ penalties, has the potential to perform the feature selection and deal with the correlation between the features simultaneously \cite{zou2005regularization}. The EN-Cox method was proposed by Noah Simon et al. \cite{simon2011regularization} where the Elastic Net penalty term shown in Table~\ref{tab:regularizers} with $0\leq\alpha\leq1$ and introduced into the log-partial likelihood function in Eq.~(\ref{equ:lpl}).
%\begin{eqnarray}
%\label{coselstic}
%\hat{\beta}_{en}&=&\min_\beta\{-\frac{2}{N}[\sum_{j=1}^N\delta_j X_j\beta-\delta_j log(\sum_{i\in R_j}e^{X_i\beta})]\nonumber\\
%&+&\lambda[\alpha\sum_{p=1}^P|\beta_p|+\frac{1}{2}(1-\alpha)\sum_{p=1}^P\beta_p^2]\}
%\end{eqnarray}
Different from Lasso-Cox, EN-Cox can select more than $N$ features if $N\leq P$. 

Kernel Elastic Net (KEN) algorithm \cite{vinzamuri2013cox}, which uses the concept of kernels, compensates for the drawbacks of the EN-Cox which is partially effective at dealing with the correlated features in survival data. In KEN-Cox, it builds a kernel similarity matrix for the feature space in order to incorporate the pairwise feature similarity into the Cox model. The regularizer used in KEN-Cox is defined as $\lambda\alpha||\beta||_1+\lambda(1-\alpha)\beta^TK\beta$, 
%\begin{equation}
%	\label{coxken}
%$\hat{\beta}_{ken}=\min_\beta LL(\beta)+\lambda\alpha\parallel\beta\parallel_1+\lambda(1-\alpha)\beta^TK\beta$
%\end{equation}
where $K$ is a symmetric kernel matrix with $K_{ij}=exp({-\parallel x_i-x_j\parallel_2^2}/{2\sigma^2}) (i,j=1,\cdots,P)$ as its entries. We can see that the equation for KEN-Cox method includes both smooth and non-smooth $\ell_1$ terms. %In \cite{bibid}, the author proposed a network-based rguarization method for high-dimensional Cox regression; it uses an $\ell_1$-penalty to induce sparsity of the regression coefficients and a quadratic Laplacian penalty to encourage smoothness between the coefficients of neighboring features on a given network. This method is named as LapNet-Cox method.\\
%@article{sun2014network,
%title={Network-regularized high-dimensional cox regression for analysis of genomic data},
%author={Sun, Hokeun and Lin, Wei and Feng, Rui and Li, Hongzhe},
%journal={Statistica Sinica},
%volume={24},
%number={3},
%pages={1433},
%year={2014},
%publisher={NIH Public Access}
%}

\vspace{2mm}
\textbf{OSCAR-Cox:} The modified graph Octagonal Shrinkage and Clustering Algorithm for Regression (OSCAR)~\cite{yang2012feature,ye2012sparse} regularizer is incorporated in the basic Cox model as the OSCAR-Cox algorithm~\cite{vinzamuri2013cox}, which can perform the variable selection for highly correlated features in regression problem. The main advantage of OSCAR regularizer is that it tends to have equal coefficients for the features which relate to the outcome in similar ways. In addition, it can simultaneously obtain the advantages of the individual sparsity because of the $\ell_1$ norm and the group sparsity due to the $\ell_\infty$ norm. The regularizer used in the formulation of the OSCAR-Cox is given in Table~\ref{tab:regularizers},
%\begin{equation}
%\label{coxoscar}
%\hat{\beta}_{ken}=\min_\beta LL(\beta)+\lambda_1\parallel\beta\parallel_1+\lambda_2T\parallel\beta\parallel_1
%\end{equation}
where $T$ is the sparse symmetric edge set matrix generated by building a graph structure which considers each feature as an individual node. By using this way, a pairwise feature regularizer can be incorporated into the basic Cox regression framework. 

%The modified Graph OSCAR (GOSCAR) regularizer with the form of $\lambda_1(\parallel\beta\parallel_1)+\lambda_2(\parallel E \beta\parallel_1)$ is used in the GOSCAR-Cox method, where $E$ is the incidence matrix of the feature graph %and $L(\beta)$ is the loss function which is the modified least squares loss function derived from the partial log likelihood of Cox regression 
%and $\lambda_1$ and $\lambda_2$ are the regularization parameters. Since the formulation in GOSCAR-COX is non-smooth, it can encourage similar coefficient values for correlated variables and also be effective at handling structured sparsity which cannot be inherently detected using the elastic net regularizers. 
%\textbf{GLasso-Cox:} In Group lasso penalized Cox regression method (GLasso-Cox) \cite{bibid}, the set of coefficients are grouped into $K$ groups for the purpose of variable selection. The formulation of the GLasso-Cox is given in Table \ref{tab:regularizers},
%\begin{equation}
%\label{coxglasso}
%\hat{\beta}_{ken}=\min_\beta LL(\beta)+\lambda \sum_{k=1}^{K}\sqrt{\sum_{j=1}^{p_k}\beta_{kj}^2}
%\end{equation}
%where $p_k$ is the number of features in the $k^{th}$ group. Group lasso performs group selection as it is singular at the group level \cite{yuan2006model}.\\

Among the regularizers shown in Table \ref{tab:regularizers}, the parameters $\lambda\geq0$ can be tuned to adjust the influence introduced by the regularizer term. The performance of these penalized estimators significantly depend on $\lambda$, and the optimal $\lambda_{opt}$ can be chosen via cross-validation. %Also, few other penalties based on kernel and graph based similarities have been recently proposed to tackle the inherent correlations within the variables in the context of Cox proportional hazards model \cite{vinzamuri2013cox}.
{The time complexity of both Lasso-Cox and EN-Cox method is $O(NP)$.}

\vspace{-1.5mm}
\subsubsection{CoxBoost}
%When predictive survival models are built from high-dimensional data, sometimes the mandatory inclusion of some additional covariates into the final model is needed. 
While there exists several algorithms (such as the penalized parameter estimation) which can be applied to fit the sparse survival models on the high-dimensional data, none of them are applicable in the situation that some mandatory covariates should be taken into consideration explicitly in the models. CoxBoost~\cite{binder2008allowing} approach is proposed to incorporate the mandatory covariates into the final model. The CoxBoost method also aims at estimating the coefficients $\beta$ in Eq.~(\ref{equ:cox}) as in the Cox model. It considers a flexible set of candidate variables for updating in each boosting step by employing the offset-based gradient boosting approach. This is the key difference from the regular gradient boosting approach, which either updates only one component of $\beta$ in component-wise boosting or fits the gradient by using all covariates in each step. %In the boosting step $k$ $(k=1,\cdots,M)$ of CoxBoost approach, $q_k$ candidate sets of covariates are predetermined with indices $I_{kl}\subset\{1,\cdots,P\}, l=1,\cdots,q_k$. For each candidate set, all the parameters for the corresponding covariates are updated simultaneously. The one which mostly improves the overall fit will be selected for the update in each step.

\subsubsection{Time-dependent (TD) Cox Model}\
\label{sec:ext-cox}
Cox regression model is also effectively adapted to handle time-dependent covariates, which refer to the variables whose values may change with time $t$ for a given instance. Typically, the time-dependent variable can be classified into three categories~\cite{kleinbaum2006survival}: internal time-dependent variable, ancillary time-dependent variable and defined time-dependent variable. The reason for a change in the internal time-dependent variable depends on the internal characteristics or behavior that is specific to the individual. In contrast, a variable is called an ancillary time-dependent variable if its value changes primarily due to the environment that may affect several individuals simultaneously. Defined variable, with the form of the product of a time-independent variable multiplied by a function of time, is used to analyze a time independent predictor not satisfying the PH assumption in the Cox model. The commonly used layout of the dataset in time-dependent Cox model is in the form of counting process (CP) \cite{kleinbaum2006survival}.

Given a survival analysis problem which involves both time-dependent and time-independent features, we can denote the variables at time $t$ as ${X}(t)=(X_{\cdot 1}(t),X_{\cdot 2}(t),...,X_{\cdot P_1}(t),X_{\cdot 1},X_{\cdot 2},...,X_{\cdot P_2})$, where $P_1$ and $P_2$ represent the number of time-dependent and time-independent variables, respectively. And $X_{\cdot j}(t)$ and $X_{\cdot i}$ represent the $j^{th}$ time-dependent feature and the $i^{th}$ time-independent feature, respectively. Then, by involving the time-dependent features into the basic Cox model given in Eq.~(\ref{equ:cox}), the time-dependent Cox model can be formulated as: % the basic Cox model shown in Eq.~(\ref{equ:cox}) can be updated to the time-dependent Cox model shown in Eq.~(\ref{Equ:tdcox}).
\vspace{-1.3mm}
\begin{equation}
\label{Equ:tdcox}
h(t,{X}(t))=h_0(t)exp\bigg[\sum_{j=1}^{P_1}\delta_j X_{\cdot j}(t)+\sum_{i=1}^{P_2}\beta_i X_{\cdot i}\bigg]
\end{equation}
where $\delta_j$ and $\beta_i$ represent the coefficients corresponding to the $j^{th}$ time-dependent variable and the $i^{th}$ time-independent variable, respectively. For the two sets of predictors at time $t$: ${X}(t)=(X_{\cdot 1}(t),X_{\cdot 2}(t),...,X_{\cdot P_1}(t),X_{\cdot 1},X_{\cdot 2},...,X_{\cdot P_2})$ and ${X}^*(t)=(X^*_{\cdot 1}(t),X^*_{\cdot 2}(t),...,X^*_{\cdot P_1}(t),X^*_{\cdot 1},X^*_{\cdot 2},...,X^*_{\cdot P_2})$, the hazard ratio for the time-dependent Cox model can be computed as follows:
\begin{equation}
\label{Equ:tdcox-hazard}
\hat{HR}(t)=\frac{\hat{h}(t,{X}^*(t))}{\hat{h}(t,{X}(t))}=exp\bigg[\sum_{j=1}^{P_1}\delta_j [X^*_{\cdot j}(t)-X_{\cdot j}(t)]+\sum_{i=1}^{P_2}\beta_i [X^*_{\cdot i}-X_{\cdot i}]\bigg]
\end{equation}
Since the first component in the exponent of~Eq. (\ref{Equ:tdcox-hazard}) is time-dependent, we can consider the hazard ratio in the TD-Cox model as a function of time $t$. This means that it does not satisfy the 
PH assumption mentioned in the standard Cox model. It should be noted that the coefficient $\delta_j$ is in itself not time-dependent and it represents the overall effect of the $j^{th}$ time-dependent variable at various survival time points. The likelihood function of time-dependent Cox model can be constructed in the same manner {and optimized with the same time complexity} as done in the Cox model.
%\vspace{-0.9mm}

\subsection{Parametric Models}
The parametric censored regression models assume that the survival times or the logarithm of the survival times of all instances in the data follow a particular theoretical distribution \cite{lee2003statistical}. These models are important alternatives to the Cox-based semi-parametric models and are also widely used in many application domains. It is simple, efficient and effective in predicting the time to event of interest using parametric methods. The parametric survival models tend to obtain the survival estimates that are consistent with a theoretical survival distribution. The commonly used distributions in parametric censored regression models are: normal, exponential, weibull, logistic, log-logistic and  log-normal. If the survival times of all instances in the data follow these distributions, the model is referred as linear regression model. If the logarithm of the survival times of all instances follow these distributions, the problem can be analyzed using the accelerated failure time model, in which we assume that the variable can affect the time to the event of interest of an instance by some constant factor \cite{lee2003statistical}. It should be noted that if no suitable theoretical distribution is known, nonparametric methods are more efficient.

The maximum-likelihood estimation (MLE) method~\cite{lee2003statistical} can be used to estimate the parameters for these models. Let us assume that the number of instances is $N$ with $c$ censored observations and $(N-c)$ uncensored observations, and use $\beta=(\beta_1,\beta_2,\cdots,\beta_P)^T$ as a general notation to denote the set of all parameters \cite{li2016regularized}. Then the death density function $f(t)$ and the survival function $S(t)$ of the survival time can be represented as $f(t,\beta)$ and $S(t,\beta)$, respectively. For a given instance $i$, if it is censored, the actual survival time will not be available. However, we can conclude that the instance $i$ did not experience the event of interest before the censoring time $C_i$, so the value of the survival function $S(C_i,\beta)$ will be a probability closed to $1$. In contrast, if the event occurs for instance $i$ at $T_i$, then the death density function $f(T_i,\beta)$ will have a high probability value. Thus, we can denote $\prod\limits_{\delta_i=1}f(T_i,\beta)$ as the joint probability of all the uncensored observations and $\prod\limits_{\delta_i=0}S(T_i,\beta)$ to represent the joint probability of the $c$ censored observations \cite{li2016regularized}. Therefore, we can estimate the parameters $\beta$ by optimizing the likelihood function of all $N$ instances in the form of %$L(\beta)=\prod\limits_{\delta_i=1}f(T_i,\beta)\prod\limits_{\delta_i=0}S(T_i,\beta)$. 
\begin{equation}
	\label{equ:parametric}
	L(\beta)=\prod\limits_{\delta_i=1}f(T_i,\beta)\prod\limits_{\delta_i=0}S(T_i,\beta)
\end{equation} 

Table \ref{tab:distributions} shows the death density function $f(t)$ and its corresponding survival function $S(t)$ and hazard function $h(t)$ for these commonly used distributions. 
%All the parametric models can be implemented using the R package called \emph{survival}. 
Now we will discuss more details about these distributions.
% Table
\begin{table}%
	\setlength\extrarowheight{7pt}		
	\tbl{Density, Survival and Hazard functions for the distributions commonly used in the parametric methods in survival analysis.\label{tab:distributions}}{%
		\centering
		\begin{tabular}{|c|c|c|c|}
			\hline
			\textbf{Distribution}&\textbf{PDF} \textbf{$f(t)$} & \textbf{Survival $S(t)$} & \textbf{Hazard $h(t)$} \\\hline	
			\hline
			Exponential &$\lambda exp(-\lambda t)$& $exp{(-\lambda t)}$ &$\lambda$\\\hline
			Weibull &$\lambda kt^{k-1}exp{(-\lambda t^k)}$ &$exp{(-\lambda t^k)}$ &$\lambda kt^{k-1}$  \\\hline
			Logistic & $\frac{e^{-(t-\mu)/\sigma}}{\sigma(1+e^{-(t-\mu)/\sigma})^2}$&$\frac{e^{-(t-\mu)/\sigma}}{1+e^{-(t-\mu)/\sigma}}$  &$\frac{1}{\sigma(1+e^{-(t-\mu)/\sigma})}$ \\\hline
			%Normal &$\frac{\lambda}{\sqrt{2\pi}}exp(-\frac{1}{2}(\lambda t)^2)$& $1-\Phi(\lambda t)$ &$\frac{\lambda}{\sqrt{2\pi}(1-\Phi(\lambda t))}exp(-\frac{1}{2}(\lambda t)^2)$\\\hline
			
			Log-logistic &$\frac{\lambda kt^{k-1}}{(1+\lambda t^k)^2}$ &$\frac{1}{1+\lambda t^k}$ &$\frac{\lambda kt^{k-1}}{1+\lambda t^k}$ \\\hline
			
			Normal &$\frac{1}{\sqrt{2\pi}\sigma}exp(-\frac{(t-\mu)^2}{2\sigma^2})$& $1-\Phi(\frac{t-\mu)}{\sigma})$ &$\frac{1}{\sqrt{2\pi}\sigma(1-\Phi((t-\mu)/\sigma))}exp(-\frac{(t-\mu)^2}{2\sigma^2})$\\\hline

			Log-normal &$\frac{1}{\sqrt{2\pi}\sigma t}exp{(-\frac{(log(t)-\mu)^2}{2\sigma^2})}$ &$1-\Phi(\frac{log(t)-\mu}{\sigma})$&$\frac{\frac{1}{\sqrt{2\pi}\sigma t}exp{(-{(log(t)-\mu)^2}/{2\sigma^2})}}{1-\Phi(\frac{log(t)-\mu}{\sigma})}$ \\\hline
			%Gamma&$\frac{\lambda^kt^{k-1}e^{-\lambda t}}{\Gamma(k)}$&$1-\frac{\int_{0}^{t}\sigma^{k-1}e^{-\sigma}d\sigma}{\Gamma(k)}$&$\frac{\lambda^kt^{k-1}e^{-\lambda t}}{(1-\frac{\int_{0}^{t}\sigma^{k-1}e^{-\sigma}d\sigma}{\Gamma(k)})\Gamma(k)}$ \\\hline						
		\end{tabular}}
		%\vspace{-0.5mm}
	\end{table}%
	
	%% Table with mean and variance
	%\begin{table}%
	%\tbl{Density, Survival and Hazard functions for commonly used distributions\label{tab:distributions}}{%
	%\begin{tabular}{|l|l|l|l|l|l|l|}
	%\hline
	%Distribution&PDF $f(t)$ & Survival $S(t)$ & Hazard $h(t)$ & Mean $E(t)$ & Variance $V(t)$ & qth-quantile $t_q$ \\\hline	
	%\hline
	%Weibull&$\lambda k(\lambda t)^{k-1}exp{(-(\lambda t)^k)}$ &$exp{((-\lambda t)^k)}$ &$\lambda kt^{k-1}$ &$\lambda^{-1}\Gamma(1+\frac{1}{k})$ &$\lambda^{-2}(\Gamma(1+\frac{2}{k})-\Gamma(1+\frac{1}{k})^2)$ &$\lambda (-log(1-q))^{1/k}$  \\\hline
	%Exponential&$\lambda exp(-\lambda t)$& $exp{(-\lambda t)}$ &$\lambda$&$\lambda^{-1}$&$\lambda^{-2}$ &\\\hline
	%Log-logistic&$\lambda pt^{p-1}(1+\lambda t^p)^{-2}$ &$\frac{1}{1+\lambda t^p}$ &$\frac{\lambda pt^{p-1}}{1+\lambda t^p}$ & & &\\\hline
	%Log-normal&$\frac{1}{\sqrt{2\pi}\sigma t}exp{(-\frac{(ln(t)-\mu)^2}{2\sigma^2})}$&$exp(\mu+\frac{\sigma^2}{2})$&$exp(2\mu+\sigma^2)(exp(\sigma^2)-1)$ & & &\\\hline
	%Gamma&$\frac{\lambda^kt^{k-1}e^{-\lambda t}}{\Gamma(k)}$&$1-\frac{\int_{0}^{t}\sigma^{k-1}e^{-\sigma}d\sigma}{\Gamma(k)}$&$\frac{\lambda^kt^{k-1}e^{-\lambda t}}{(1-\frac{\int_{0}^{t}\sigma^{k-1}e^{-\sigma}d\sigma}{\Gamma(k)})\Gamma(k)}$ & & &\\\hline			
	%\end{tabular}}
	%\end{table}%
	
	%\begin{figure*}[h]
	%	\centering
	%	%\scalebox{.5}{\includegraphics{example}}
	%	\scalebox{.35}{\includegraphics[trim={0.2cm 0.2cm 0.2cm 0.2cm},clip]{figure/distribution}} \hspace{-0.1 cm}
	%	\caption{distribution.}
	%	\label{Fig:distribution}
	%\end{figure*}
\vspace{2mm}	
\textbf{Exponential Distribution:} Among the parametric models in survival analysis, exponential model is the simplest and prominent one since it is characterized by a constant hazard rate, $\lambda$, which is the only parameter. In this case, the failure or the death is assumed to be a random event independent of time. A larger value of $\lambda$ indicates a higher risk and a shorter survival time period. Based on the survival function shown in Table \ref{tab:distributions}, we can have $logS(t)=-\lambda t$ 
	%\begin{equation}
	%\nonumber
	%\label{exponential}
	%logS(t)=-\lambda t
	%\end{equation}
, in which the relationship between the logarithm of survival function and time $t$ is linear with $\lambda$ as the slope. Thus, it is easy to determine whether the time follows an exponential distribution by plotting $log\hat{S}(t)$ against time $t$~\cite{lee2003statistical}.
	
	%However, the assumption that the hazard function is a constant for each pattern of covariates is a much stronger assumption than the PH assumption. If the hazard function is constant, then the hazard ratio will be constant. On the other hand, the hazard ratio being constant does not necessarily mean that each hazard is constant. In other words, the baseline hazard function in Cox model is not specified, but it is a PH model. It should be noted that, parametric survival models need not be PH models. Many parametric models are acceleration failure (AFT) models rather than PH models. The exponential and Weibull distributions can accommodate both the PH and AFT assumptions. For the interpretation of the estimated results, it also differs for AFT models and PH models. While the comparison of hazards is the goal of the PH based models, the AFT models allow for directly comparing the survival times. %It aims to compare the hazards for PH model, while AFT is applicable for a comparison of survival times.

\vspace{2mm}
\textbf{Weibull Distribution:} The Weibull model, which is characterized by two parameters $\lambda>0$ and $k>0$, is the most widely used parametric distribution for survival problem. The shape of the hazard function is determined using the shape parameter $k$, which provides more flexibility compared to the exponential model. If $k=1$, the hazard will be a constant, and in this case, the Weibull model will become an exponential model. %with $h(t)=\lambda$. 
If $k<1$, the hazard function will be decreasing over time. The scaling of the hazard function is determined by the scaling parameter $\lambda$.

%The two important properties of the Weibull model are: (i) If the AFT assumption holds, then the PH assumption in Cox model also holds, (ii) Based on the survival function $S(t)=exp((-\lambda t)^k)$, we can have $log[-logS(t)]=k\cdot log(\lambda)+k\cdot log(t)$ which means that 
%%\begin{equation}
%%\nonumber
%%\label{weibull_linear}
%%log[-logS(t)]=k\cdot log(\lambda)+k\cdot log(t)
%%\end{equation}
%%It means that 
%for Weibull model $log[-logS(t)]$ is a linear function of $log(t)$ with slope $k$ and intercept $k\cdot log(\lambda)$. %By this key property, we can also evaluate the Weibull model by plot this linear relationship.

\vspace{2mm}	
\textbf{Logistic and Log-logistic Distribution:} %Log-logistic distribution model accommodates an AFT model but not a PH model. 
In contrast to Weibull model, the hazard functions of both logistic and log-logistic models allow for non-monotonic behavior in the hazard function, which is shown in Table \ref{tab:distributions}.
	%\begin{equation}
	%\label{loglogistic}
	%h(t)=\frac{\lambda pt^{p-1}}{1+\lambda t^p}
	%\end{equation}
The survival time $T$ and the logarithm of survival time $log(T)$ will follow the logistic distribution in logistic and log-logistic models, respectively. For logistic model, $\mu$ is the parameter to determine the location of the function, while $\sigma$ is the scale parameter. For log-logistic model, the parameter $k>0$ is the shape parameter. If $k\le 1$, the hazard function is decreasing over time. However, if $k>1$, the hazard function will increase over time to the maximum value first and then decrease, which means that the hazard function is unimodal if $k>1$. Thus, the log-logistic distribution may be used to describe a monotonically decreasing hazard or a first increasing and then decreasing hazard \cite{lee2003statistical}. 
	
	%The Log-logistic AFT model is a proportional odds (PO) model \cite{kleinbaum2006survival} and not a PH model. For a PO survival model, the odds ratio is assumed to remain constant over time. 
	%
	%Here, we introduce other definitions for Log-logistic model that are available in the literature.
	%The survival odds (SO) \cite{kleinbaum2006survival} is defined as the odds of surviving beyond time $t$ in the form of 
	%\begin{equation}
	%%\nonumber
	%\label{survodds}
	%\frac{S(t)}{1-S(t)}=\frac{P(T> t)}{P(T\le t)}
	%\end{equation}
	%and the failure odds (FO) \cite{kleinbaum2006survival}, which is the reciprocal of survival odds, which means the odds of getting the event by time $t$ in the form of 
	%\begin{equation}
	%%\nonumber
	%\label{failodds}
	%\frac{1-S(t)}{S(t)}=\frac{P(T\le t)}{P(T>t)}
	%\end{equation}
	%According to the Log-logistic survival function in Table \ref{tab:distributions}, the failure odds equals to $\lambda t^k$, which indicates a linear relationship between the log of failure odds  and the log of time. This is also helpful for the evaluation by plotting $log(FO)$ against $log(t)$. The plots will be a line with slop $k$ represented by $log(FO)=ln(\lambda)+k\cdot log(t)$ if the survival time in the data follows a Log-logistic distribution.
	%%\begin{equation}
	%%\nonumber
	%%\label{failodds_new}
	%%log(FO)=ln(\lambda)+k\cdot log(t)
	%%\end{equation}

\vspace{2mm}	
\textbf{Normal and Log-normal Distribution:} If the survival time $T$ satisfies the condition that  $T$ or $log(T)$ is normally distributed with mean $\mu$ and variance $\sigma^2$, then $T$ is normally or log-normally distributed. This is suitable for the survival patterns with an initially increasing and then decreasing hazard rate. %The Log-normal AFT model does not belong to the PH model. %The Log-normal model has a relatively complicated hazard and survival function that can only be expressed in terms of integrals. \\
	
	%\textbf{Generalized Gamma Distribution} The generalized Gamma Distribution, characterized by three parameters, $\lambda$, $\alpha$ and $\gamma$, is defined with the density function 
	%\begin{equation}
	%\nonumber
	%\label{gamma}
	%f(t)=\alpha \lambda^{\alpha \lambda}t^{\alpha\gamma-1} exp (-(\lambda t)^\alpha)/\Gamma(\gamma)
	%\end{equation}
	%It's easy to find that the Exponential, the Weibull, the Log-nomal and the Gamma distributions are special cases of the generalized gamma distribution, when $\alpha=\gamma=1$, $\gamma=1$, $\gamma \rightarrow \infty$ and $\alpha=1$, respectively.

	%The hazard and survival function of generalized Gamma model is also complicated. It has three parameters allowing more flexibility in its shape. Actually, the Weibull and Log-normal distributions are two special cases of the generalized gamma distribution. 
	
	%\subsection{Tobit Model} Tobit model is an extension of linear regression $y_j=X_j\beta+\epsilon_j, \epsilon_j\sim N(0, \sigma^2)$, but the parameter is estimated by the maximum likelihood method rather than least square error. It uses the parametric method framework with the probability density function and the cumulative distribution function of the standard normal distribution.
	
Based on the framework given in Eq. (\ref{equ:parametric}), we will discuss these commonly used parametric methods. 
%\vspace{-1.3mm}
\subsubsection{Linear regression models}
In data analysis, the linear regression model, together with the least squares estimation method, is one of the most commonly used approach. We cannot apply it directly to solve survival analysis problems since the actual event times are missing for censored instances. Some linear models~\cite{miller1982regression,koul1981regression,buckley1979linear,wang2008doubly,li2016regularized} including Tobit regression and Buckley-James (BJ) regression were proposed to handle censored instances in survival analysis. Strictly speaking, linear regression is a specific parametric censored regression, however, this method is fundamental in data analysis, and hence we discuss the linear regression methods for censored data separately here.

\vspace{2mm}
\textbf{Tobit Regression:}
The Tobit model~\cite{tobin1958estimation} is one of the earliest attempts to extend linear regression with the Gaussian distribution for data analysis with censored observations. In this model, a latent variable $y^*$ is introduced and the assumption made here is that it linearly depends on $X$ via the parameter $\beta$ as $y^*=X\beta +\epsilon, \epsilon\sim N(0,\sigma^2)$, where $\epsilon$ is a normally distributed error term. Then, for the $i^{th}$ instance, the observable variable $y_i$ will be $y_i^*$ if $y_i^* > 0$, otherwise it will be $0$.
%\begin{equation}
%\nonumber
%y_i=\begin{cases}
%y_i^* &\text{if }  y_i^* > 0 \\
%0 &\text{if } y_i^* \le 0
%\end{cases}
%\end{equation}
This means that if the latent variable is above zero, the observed variable equals to the latent variable and zero otherwise. Based on the latent variable, the parameters in the model can be estimated with maximum likelihood estimation (MLE) method {with time complexity $O(NP^2)$}.

\vspace{2mm}
\textbf{Buckley-James Regression:}
The Buckley-James (BJ) regression~\cite{buckley1979linear} estimates the survival time of the censored instances as the response value based on Kaplan-Meier (KM) estimation method, and then fits a linear (AFT) model by considering the survival times of uncensored instances and the approximated survival times of the censored instances at the same time. %The authors in \cite{miller1976least} suggest estimate the parameter in the linear model by minimizing
%\begin{equation}
%	\int \epsilon^2 d\hat{F}_{\beta}(\epsilon)
%\end{equation}
%where $\hat{F}_{\beta}(\epsilon)=1-\prod\limits_{i:e_i\le \delta_i}(\frac{N-i}{N-i+1})^{\delta_i}$ is the non-parametric estimator of $F$ based on the censored and uncensored residuals $e_i(\beta)=y_i-X_i\beta\ (i=1,\cdots, N)$. 
To handle high-dimensional survival data, Wang et al.~\cite{wang2008doubly} applied the elastic net regularizer in the BJ regression (EN-BJ).

\vspace{2mm}
\textbf{Penalized Regression:} 
%One of the critical challenges with modeling such survival data is the presence of censored outcomes which cannot be handled by standard regression models. 
Penalized regression methods~\cite{kyung2010penalized} are well-known for their nice properties of simultaneous variable selection and coefficient estimation. The penalized regression method can provide better prediction results in the presence of either multi-collinearity of the covariates or high-dimensionality. Recently, these methods have received a great attention in survival analysis. The weighted linear regression model with different regularizers for high-dimensional censored data is an efficient method to handle the censored data by giving different weights to different instances \cite{li2016regularized}. In addition, the structured regularization based linear regression algorithm \cite{bach2012structured,bhanu2016} for right censored data has a good ability to infer the underlying structure of the survival data.  

\begin{itemize}
%\vspace{-2mm}
\item \textbf{Weighted Regression:}
Weighted regression method~\cite{li2016weighted} can be used when the constant variance assumption about the errors in the ordinary least squares regression methods is violated (which is called heteroscedasticity), which is different from the constant variance in the errors (which is called homoscedasticity) in ordinary least squares regression methods. Instead of minimizing the residual sum of squares, the weighted regression method minimizes the weighted sum of squares  $\sum_{i=1}^{n}w_i(y_i-X_i\beta)^2$. %as follows
%\begin{equation}
%WSS(\beta, w)=\sum_{i=1}^{n}w_i(y_i-x_i\cdot\beta)^2
%\end{equation}
The ordinary least squares is a special case of this where all the weights $w_i=1$. Weighted regression method can be solved in the same manner as the ordinary linear least squares problem {with the time complexity of $O(NP)$}. In addition, using the weighted regression method, we can assign higher weights to the instances that we want to emphasize or ones where mistakes are especially costly. If we give the samples high weights, the model will be pulled towards matching the data. This will be very helpful for survival analysis to put more emphasis on the instances whose information may contribute more to the model. %Typically, the reciprocal of the variance of the error is assigned as the weights. 
\vspace{2mm}
\item \textbf{Structured Regularization:}
The ability to effectively infer latent knowledge through tree-based hierarchies and graph-based relationships is extremely crucial in survival analysis. This is also supported by the effectiveness of structured sparsity based regularization methods in regression \cite{bach2012structured}. Structured regularization based LInear REgression algorithm for right Censored data (SLIREC) in \cite{bhanu2016} infers the underlying structure of the survival data directly using sparse inverse covariance estimation (SICE) method and uses the structural knowledge to guide the base linear regression model. %There are two stages in SLIREC algorithm:
The structured approach is more robust compared to the standard statistical and Cox based methods since it can automatically adapt to different distributions of events and censored instances.% which is very useful when dealing with different real-world datasets.
\end{itemize}
%\vspace{-4mm}
\subsubsection{Accelerated Failure Time (AFT) Model}
In the parametric censored regression methods discussed previously, we assume that the survival time of all instances in the given data follows a specific distribution and that the relationship between either the survival time or the logarithm of the survival time and the features is linear. Specially, if the relationship between the logarithm of survival time $T$ and the covariates is linear in nature, it is also termed as Accelerated failure time (AFT) model \cite{kalbfleisch2011statistical}. Thus, we consider these regression methods as the generalized linear models. 

%Based on the representation of a triple of variables $(X,Z,\delta)$ for each of the individual, consider the $i^{th}$ individual with the covariate vector$X_i=(x_{i1},x_{i2},\cdots,x_{iP})$, time $Z_i$ and coefficient vector $\beta^T=(\beta_1,\beta_2,...,\beta_P)$.\\

In the AFT model, it assumes that the relationship of the logarithm of survival time $T$ and the covariates is linear and can be written in the following form.
\begin{equation}
\label{linear}
ln(T)=X\beta+\sigma \epsilon
\end{equation}
where $X$ is the covariate matrix, $\beta$ represents the coefficient vector, $\sigma(\sigma>0)$ denotes an unknown scale parameter, and $\epsilon$ is an error variable which follows a similar distribution to $ln(T)$. Typically, we make a parametric assumption on $\epsilon$ which can follow any of the distributions given in Table~\ref{tab:distributions}. In this case, the survival is dependent on both the covariate and the underlying distribution. Then, the only distinction of an AFT model compared to regular linear methods would be the inclusion of censored information in the survival analysis problem. The AFT model is additive with respect to $ln(T)$, while multiplicative with respect to $T$, and is written in the form of $T=e^{X\beta}e^{\sigma \epsilon}$.
%\begin{equation}
%\label{multiplicative}
%T=e^{X\beta}e^{\sigma \epsilon}
%\end{equation}

%Once the probability density function is specified for survival time, the corresponding survival and hazard functions can be determined using Equation in section 2. Actually, for the probability density function, suvival function and hazard function, we can ascertain other two functions by specifying any one of them.
{Thus, AFT model assumes that the features have the multiplicative effect on the survival time.} In order to demonstrate {the basic function of this assumption on a specific feature}, let us compare the survival {functions $S_1(t)$ and $S_2(t)$ for two groups, where the instances in the same group have the same values on this feature but different values if they belong to different groups}. Then, the assumption of AFT model is in the form of 
\begin{equation}
\label{aft}
S_2(t)=S_1(\gamma t)
\end{equation}
where $t \ge 0$ and $\gamma$ represents a constant which is named as an acceleration factor for comparison of the {survival time of the two groups}. For the linear regression method, we can parameterize $\gamma$ as $exp(\alpha)$, where $\alpha$ can be estimated using the given data. Then, the assumption in AFT method will be updated to $S_2(t)=S_1(exp(\alpha)t)$.
%\begin{equation}
%\label{aft}
%S_2(t)=S_1(exp(\alpha) t) 
%\end{equation}
The acceleration factor which is the key measure of the relationships in the AFT method is used to evaluate the effect of features on the survival time. %Table \ref{tab:distributions} shows the death density function $f(t)$ and its corresponding survival function $S(t)$ and hazard function $h(t)$ for these commonly used distributions in AFT models. 
{The time complexity of AFT models is also $O(NP^2)$.}

%\vspace{-2mm} 
\section{Machine Learning methods}
\label{sec4}
In the past several years, due to the advantages of machine learning techniques, such as its ability to model the non-linear relationships and the quality of their overall predictions made, they have achieved significant success in various practical domains. In survival analysis, the main challenge of machine learning methods is the difficulty to appropriately deal with censored information and the time estimation of the model. Machine learning is effective when there are a large number of instances in a reasonable dimensional feature space, but this is not the case for certain problems in survival analysis~\cite{zupan2000machine}. In this section, we will do a comprehensive review of commonly used machine learning methods in survival analysis.
%\vspace{-3mm} 
\subsection{Survival Trees}
Survival trees are one form of classification and regression trees which are tailored to handle censored data. The basic intuition behind the tree models is to recursively partition the data based on a particular splitting criterion, and the objects that are similar to each other based on the event of interest will be placed in the same node. The earliest attempt at using a tree structure for survival data was made in~\cite{ciampi1981approach}. However, \cite{gordon1985tree} is the first paper which discussed the creation of survival trees.

The primary difference between a survival tree and the standard decision tree is in the choice of splitting criterion. The decision tree method performs recursive partitioning on the data by setting a threshold for each feature, however, it can neither consider the interactions between the features nor the censored information in the model~\cite{safavian1991survey}. The splitting criteria used for survival trees can be grouped into two categories: (i) maximizing between-node heterogeneity and (ii) minimizing within-node homogeneity. The first class of approaches minimizes the loss function using the within-node homogeneity criterion. The authors in \cite{gordon1985tree} measured the homogeneity and Hellinger distances between the estimated distribution functions using the Wasserstein metric . An exponential log-likelihood function was employed in~\cite{davis1989exponential} for recursive partitioning based on the sum of residuals from the Cox model. Leblanc and Crowley \cite{leblanc1992relative} measured the node deviance based on the first step of a full likelihood estimation procedure. In the second class of splitting criteria, Ciampi et al. \cite{ciampi1986stratification} employed log-rank test statistics for between-node heterogeneity measures. Later, Ciampi et al. \cite{ciampi1987recursive} proposed a likelihood ratio statistic to measure the dissimilarity between two nodes. Based on the Tarone-Ware class of two-sample statistics, Segal \cite{segal1988regression} introduced a procedure to measure the between-node dissimilarity. The main improvement of a survival tree over the standard decision tree is its ability to handle the censored data using the tree structure.

Another important aspect of building a survival tree is the selection of the final tree. Procedures such as backward selection or forward selection can be followed for choosing the optimal tree~\cite{bou2011review}. However, an ensemble of trees (described in Section \ref{sec40}) can avoid the problem of final tree selection with better performance compared to a single tree.
\vspace{-1mm}
\subsection{Bayesian Methods}
Bayes theorem is one of the most fundamental principles in probability theory and mathematical statistics; it provides a link between the \emph{posterior probability} and the \emph{prior probability}, so that one can see the changes in probability values before and after accounting for a certain event. %The formulation of the Bayes theorem is
%\begin{equation}
%\label{equ:BT}
%Pr(Y|X)=\frac{Pr(X|Y)\cdot Pr(Y)}{Pr(X)},
%\end{equation}
%where $Pr(Y|X)$ is the probability of event $Y$, conditional upon event $X$.
Using the Bayes theorem, there are two models, namely, Na\"{i}ve Bayes (NB) and Bayesian network (BN) \cite{friedman1997bayesian}. Both of these approaches, which provide the probability of the event of interests as their outputs, are commonly studied in the context of clinical prediction \cite{kononenko1993inductive,pepe2003statistical,zupan2000machine}. The experimental results of using Bayesian methods on survival data show that Bayesian methods have good properties of both interpretability and uncertainty reasoning~\cite{Raftery95accountingfor}.

{Na\"{i}ve Bayes}, a well-known probabilistic method in machine learning, is one of the simplest yet effective prediction algorithms. In~\cite{bellazzi2008predictive}, the authors build a na\"{i}ve Bayesian classifier to make predictions in clinical medicine by estimating various probabilities from the data. Recently, the authors in \cite{early2016} effectively integrate Bayesian methods with an AFT model by extrapolating the prior event probability to implement early stage prediction on survival data for the future time points. %More specifically, they developed two probabilistic algorithms based on Naive Bayes and Tree-Augmented Naive Bayes (TAN), called ESP-NB and ESP-TAN, respectively, for early stage event prediction by modifying the posterior probability of event occurrence using different extrapolations that are based on Weibull and Lognormal distributions.
One drawback of Na\"{i}ve Bayes method is that it makes the independence assumption between all the features, which may not be true for many problems in survival analysis.

A Bayesian network, in which the features can be related to each other at various levels, can graphically represent a theoretical distribution over a set of variables. Bayesian networks can visually represent all the relationships between the variables which makes it interpretable for the end user. It can acquire knowledge information by using procedures of estimating the network structures and parameters from a given dataset. In~\cite{lisboa2003bayesian}, the authors proposed a Bayesian neural network framework to perform model selection for {survival} data using automatic relevance determination~\cite{mackay1995probable}. In~\cite{raftery1995bayesian}, a Bayesian model averaging for Cox proportional hazards models is proposed and also used to evaluate the Bayes factors in the problem. More recently, in~~\cite{early2016}, the authors proposed a novel framework which combines the power of Bayesian network representation with the AFT model by extrapolating the prior probabilities to future time points. {The time complexity of these Bayesian approaches mainly depends on the types of Bayesian method used in the models.}
\vspace{-1mm}
\subsection{Artificial Neural Networks}
Inspired by biological neural systems, in 1958, Frank Rosenblatt published the first paper~\cite{rosenblatt1958perceptron} about artificial neural network (ANN). In this approach, the simple artificial nodes denoted by ``neurons" are connected based on a weighted link to form a network which simulates a biological neural network. A neuron in this context is a computing element which consists of sets of adaptive weights and generates the output based on a certain kind of \emph{activation function}. 
%A simple artificial neural network named \emph{perceptron}, which only has input and output layers.
Artificial neural network (ANN) has been widely used in survival analysis. Three kinds of methods are proposed in the literature which employ the neural network method to solve the survival analysis problems.
\begin{enumerate}
\item The neural network survival analysis has been employed to predict the survival time of a subject directly from the given inputs. %(from the thesis)

\item The authors in~\cite{faraggi1995neural} extended the Cox PH model to the non-linear ANN predictor and suggested to fit the neural network which has a linear output layer and a single logistic hidden layer. The authors in~\cite{mariani1997prognostic} used both the standard Cox model and the neural network method proposed in~\cite{faraggi1995neural} to assess the prognostic factors for the recurrence of breast cancer. 
%The results indicated that the ANN approach showed the potential to outperform conventional regression techniques when complex interactions or non-linear effects of continuous predictors are presented in the data. 
Although these extensions for Cox model allowed for preserving most of the advantages of a typical PH model, they were still not the optimal way to model the baseline variation~\cite{baesens2005neural}.

\item  Many approaches~\cite{liestbl1994survival,biganzoli1998feed,brown1997use,ravdin1992practical,lisboa2003bayesian} take the survival status of a subject, which can be represented by the survival or hazard probability, as the output of the neural network. The authors in~\cite{biganzoli1998feed} apply the partial logistic artificial neural network (PLANN) method to analyze the relationship between the features and the survival times in order to obtain a better predictability of the model. Recently, feed-forward neural networks are used to obtain a more flexible non-linear model by considering the censored information in the data using a generalization of both continuous and discrete time models~\cite{biganzoli1998feed}. In~\cite{lisboa2003bayesian}, the PLANN was extended to a Bayesian neural framework with covariate-specific regularization to carry model selection using automatic relevance determination \cite{mackay1995probable}.
\end{enumerate}
%\vspace{-1mm}
\subsection{Support Vector Machines}
Support Vector Machines (SVM), a very successful supervised learning approach, is used mostly for classification and can also be modified for regression problems~\cite{smola2004tutorial}. It has also been successfully adapted to in survival analysis problems.

A naive way is to consider only those instances which have events in support vector regression (SVR), in which the $\epsilon$-insensitive loss function, $f(X_i)=max(0,|f(X_i)-y_i|-\epsilon)$, is minimized with a regularizer~\cite{smola1998learning}. 
%Figure~\ref{figure:svr}~(\subref{fig:svr1}) shows the $\epsilon$-insensitive loss function, which is zero when the absolute difference between the actual and the predicted values is less than $\epsilon-$, otherwise, there is a cost which grows linearly. 
However, the main disadvantage of this approach is that the order information included in the censored instances will be completely ignored \cite{shivaswamy2007support}. Another possible approach to handle the censored data is to use support vector classification using the constraint classification approach \cite{har2002constraint} which imposes constraints in the SVM formulation for two comparable instances in order to maintain the required order. However, the computational complexity for this algorithm is quadratic with respect to the number of instances. In addition, it only focuses on the ordering among the instances, and ignores the actual values of the output.

The authors in \cite{khan2008support} proposed support vector regression for censored data (SVRc), which takes advantage of the standard SVR and also adapts it for censored cases by using an updated asymmetric loss function. In this case, it considers both the uncensored and censored instances in the model. The work in \cite{van2007support} studies a learning machine designed for predictive modeling of independently right censored survival data by introducing a health index which serves as a proxy between the instance's covariates and the outcome. The authors in~\cite{van2011support} introduces a SVR based approach which combines the ranking and regression methods in the context of survival analysis.{ In average, the time complexity of these methods is $O(N^3)$ which follows the time complexity in the standard SVM.}

Relevance Vector Machine (RVM)~\cite{widodo2011application,kiaee2016relevance}, which obtains the parsimonious estimations for regression and probabilistic problems using Bayesian inference, has the same formulation as SVM but provides probabilistic classification. RVM adopts a Bayesian approach by considering the prior over the weights controlled by some parameters. Each of these parameters corresponds to a weight, the most probable value of which can be estimated iteratively using the data.
%RVM adopts a Bayesian approach to learning, in which a prior over the weights governed by a set of hyperparameters is introduced, one associated with each weight, whose most probable value is iteratively estimated from the data. 
The Bayesian representation of the RVM can avoid these parameters in SVM (the optimization methods based on  cross-validation are usually used.). However, it is possible that RVMs converge to the local minimum since EM algorithm is used to learn the parameters. This is different from the regular sequential minimal optimization (SMO) algorithm used in SVM, which can guarantee the convergence to a global minimum.

\subsection{Advanced Machine Learning Approaches}
\label{sec40}
Over the past few years, more advanced machine learning methods have been developed to deal with and predict from censored data. These methods have various unique advantages on survival data compared to the other methods described so far.

\subsubsection{Ensemble Learning}
Ensemble learning methods~\cite{dietterich2000ensemble} generate a committee of classifiers and then predict the class labels for the new coming data points by taking a weighted vote among the prediction results from all these classifiers. It is often possible to construct good ensembles and obtain a better approximation of the unknown function by varying the initial points, especially in the presence of insufficient data. To overcome the instability of a single method, bagging \cite{breiman1996bagging} and random forests \cite{breiman2001random}, proposed by Breiman, are commonly used to perform the ensemble based model building. Such ensemble models have been successfully adapted to survival analysis {whose time complexity mainly follows that of the base-learners}. 

%For survival data, Hothorn \cite{hothorn2004bagging} proposed a general bagging method %which was implemented in the R package ``\emph{ipred}".
%and in 2008, Ishwaran introduced a general random forest method, called random survival forest (RSF) \cite{ishwaran2008random}.% and implemented it in the R package ``\emph{randomSurvivalForest}".

\vspace{2mm}
\textbf{Bagging Survival Trees:}
Bagging is one of the oldest and most commonly used ensemble method which typically reduces the variance of the base models that are used. In bagging survival trees, the aggregated survival function can be calculated by averaging the predictions made by a single survival tree instead of taking a majority vote~\cite{hothorn2004bagging}. There are mainly three steps in this method: (i) Draw $B$ booststrap samples from the given data. (ii) For each bootstrap sample, build a survival tree and ensure that, for all the terminal nodes, the number of events is greater than or equal to the threshold $d$. (iii) By averaging the leaf nodes' predictions, calculate the bootstrap aggregated survival function.
%\begin{enumerate}
%	\item Draw B booststrap samples from the original dataset.
%	\item Grow a survival tree for each bootstrap sample, and ensure that in each terminal node the number events occurred is no less than $d$.
%	\item Compute the bootstrap aggregated survival function by averaging the leaf nodes' predictions.
%\end{enumerate}
For each leaf node the survival function is estimated using the KM estimator, and all the individuals within the same node are assumed to have the same survival function.

\vspace{2mm}
\textbf{Random Survival Forests:}
Random forest is an ensemble method specifically proposed to make predictions using the tree structured models \cite{breiman2001random}. It is based on a framework similar to Bagging; the main difference between random forest and bagging is that, at a certain node, rather than using all the attributes, random forest only uses a random subset of the residual attributes to select the attributes based on the splitting criterion. It is shown that randomization can reduce the correlation among the trees and thus improve the prediction performance.

Random survival forest (RSF)~\cite{ishwaran2008random} extended Breiman's random forest method by using a forest of survival trees for prediction. There are mainly four steps in RSF: (i) Draw $B$ bootstrap samples randomly from the given dataset. This is also called out-of-bag (OOB) data because around 37\% of the data is excluded in each sample. (ii) For each sample, build a survival tree  by randomly selecting features and split the node using the candidate feature which can maximize the survival difference between the child nodes. (iii) Build the tree to the full size with a constraint that the terminal node has greater than or equal to a specific unique deaths. (iv) Using the non-parametric Nelson-Aalen estimator, calculate the ensemble cumulative hazard function (CHF) of OOB data by taking the average of the CHF of each tree.	
%\begin{enumerate}
%	\item Draw $B$ bootstrap samples randomly from the original data. This is also called out-of-bag (OOB) data because 37\% of the data is excluded in each sample.
%	\item Grow a survival tree for each sample by randomly selecting $p$ candidate variables. The node is split using the candidate variable that maximizes the survival difference between child nodes.
%	\item Grow the tree to the full size under the constraint that a terminal node has no less than $d_0$ unique deaths.
%	\item Based on the Nelson-Aalen estimator for cumulative hazard function (CHF), calculate the ensemble CHF of OOB data by taking the average of the CHF of each tree.	
%\end{enumerate}
In addition, the authors in \cite{ishwaran2011random} provide an effective way to apply RSF for high-dimensional survival analysis problems by regularizing forests.

\vspace{2mm}
\textbf{Boosting:}
Boosting algorithm is one of the widely used ensemble methods designed to combine base learners into a weighted sum that represents the final output of the strong learner. It iteratively fits the appropriately defined residuals based on the gradient descent algorithm~\cite{hothorn2006survival,buhlmann2007boosting}. The authors in \cite{hothorn2006survival} extend the gradient boosting algorithm to minimize the weighted risk function $\hat{\beta}_{\tilde{U},X}=arg\ min_\beta \sum\limits_{i=1}^N w_i(\tilde{U}_i-h(X_i|\beta))$, where $\tilde{U}$ is a pseudo-response variable with $\tilde{U}_i=-\frac{\partial L(y_i,\phi)}{\partial \phi}|_{\phi=\hat{f}_m(X_i)}$; $\beta$ is a vector of parameters; $h(\cdot|\beta_{U,X})$ is the prediction made by regressing $U$ using a base learner. Then the steps to optimize this problem are as follows: (i) Initialize $\tilde{U}_i=y_i\ (i=1,\cdots,N)$, $m=0$ and $\hat{f}_0(\cdot|\hat{\beta}_{\tilde{U},X})$; fix the number of iterations $M (M>1)$. (ii) 
%using $\tilde{U}_i=-\frac{\partial L(y_i,\phi)}{\partial \phi}|_{\phi=\hat{f}_m(X_i)}$ 
Fit $h(\cdot|\hat{\beta}_{U,X})$ after updating residuals $\tilde{U}_i\ (i=1,\cdots,N)$. (iii) Iteratively update $\hat{f}_{m+1}(\cdot)=\hat{f}_{m}(\cdot)+v h(\cdot|\hat{\beta}_{U,X})$, where $0<v\le 1$ represents the step size. (iv) Repeat the procedures in steps (ii) and (iii) until $m=M$.
%\begin{enumerate}
%	\item Initialize $\tilde{U}_i=y_i\ (i=1,\cdots,N)$; set $m=0$ and $\hat{f}_0(\cdot|\hat{\beta}_{\tilde{U},X})$; fix the number of iterations $M (M>1)$.
%	\item Compute the residuals $\tilde{U}_i\ (i=1,\cdots,N)$
%	%using $\tilde{U}_i=-\frac{\partial L(y_i,\phi)}{\partial \phi}|_{\phi=\hat{f}_m(X_i)}$ 
%	and fit the base learner $h(\cdot|\hat{\beta}_{U,X})$ to the new responses $\tilde{U}_i$ using weighted least squares.
%	\item Update $\hat{f}_{m+1}(\cdot)=\hat{f}_{m}(\cdot)+v h(\cdot|\hat{\beta}_{U,X})$ with step size $0<v\le 1$.
%	\item Increase $m$ and repeat Step 2 and 3 until $m=M$.
%\end{enumerate}
%\vspace{-3mm}
\subsubsection{Active Learning}
Active learning based on the data with censored observations can be very helpful for survival analysis since the opinions of an expert in the domain can be incorporated into the models. Active learning mechanism allows the survival model to select a subset of subjects by learning from a limited set of labeled subjects first and then query the expert to get the label of survival status before considering it in the training set. The feedback from the expert is particularly useful for improving the model in many real-world application domains~\cite{vinzamuri2014active}. %This means that the expert can integrate domain knowledge into the survival model to build a more robust model by active learning.
%In survival data, the time to the event occurrence is not necessarily observed for all the instances in the study and hence the outcome variable might be incomplete. Building models in the presence of censored data is a challenging task which has significant practical value in survival analysis. In general censored data mining tasks, censored instances are either deleted or the missing values are imputed to convert it into an uncensored problem. However, these techniques will lead to a suboptimal model due to neglecting the available information or proved an underestimate of the true performance of the model. 
The goal of active learning for survival analysis problems is to build a survival regression model by utilizing the censored instances completely without deleting or modifying the instance. In~\cite{vinzamuri2014active}, the active regularized Cox regression (ARC) algorithm based on a discriminative gradient sampling strategy is proposed by integrating the active learning method with the Cox model. The ARC framework is an iteration based algorithm with three main steps: (i) Build a regularized Cox regression using the training data, (ii) Apply the model obtained in (i) to all the instances in the unlabeled pool, (iii) Update the training data and the unlabeled pool, select the instance whose influence on the model is the highest and label it before running the next iteration. One of the main advantages of the ARC framework is that it can identify the instances and get the feedback about event labeling from the domain expert. 
{The time complexity of the ARC algorithm is $O(NPK)$, where $K$ represents the number of unique time points in the survival problem.} %$The term $NK$ is introduced due to the sampling procedure on the pool of unlabeled instances.}

%The model chooses an instance from both censored and uncensored set of instances in the dataset and query the expert to obtain the time-to-event label. 
%\vspace{-3mm}
\subsubsection{Transfer {Learning}}
%Survival analysis is an important branch of statistics which aims at predicting the time to the event of interest, and it can simultaneously model event data and censored data. 
Collecting labeled information in survival problems is very time consuming, i.e., one has to wait for the event occurrence from a sufficient number of training instances to build robust models. %Moreover, in many practical applications, appropriate feature collection can also be extremely expensive and tedious. 
A naive solution for this insufficient data problem is to merely integrate the data from related tasks into a consolidated form and build prediction models on such integrated data. However, such approaches often do not perform well because the target task (for which the predictions are to be made) will be overwhelmed by auxiliary data with different distributions. In such scenarios, knowledge transfer between related tasks will usually produce much better results compared to a data integration approach. Transfer learning method has been extensively studied to solve standard regression and classification problems \cite{pan2010survey}. %However, there are not many works which deal with transfer learning in survival analysis.
Recently, in \cite{li2016coxtransfer}, a regularized Cox PH model named Transfer-Cox, is proposed to improve the prediction performance of the Cox model in the target domain through knowledge transfer from the source domain in the context of survival models built on multiple high-dimensional datasets. The Transfer-Cox model employs $\ell_{2,1}$-norm to penalize the sum of the loss functions (negative partial log-likelihood) for both source and target domains. Thus, the model{, with time complexity $O(NP)$,} will not only select important features but will also learn a shared representation across source and target domains to improve the model performance on the target task.

\subsubsection{Multi-task Learning}
In \cite{limulti}, the survival time prediction problem is reformulated as a multi-task learning problem. In survival data, the outcome labeling matrix is incomplete since the event label of each censored instance is unavailable after its corresponding censoring time; therefore, it is not suitable to handle the censored information using the standard multi-task learning methods. To solve this problem, the multi-task learning model for survival analysis (MTLSA) translates the original event labels into a $N\times K$ indicator matrix $I$, where $K=max(y_i)\ (\forall i=1,\cdots,N)$ is the maximum follow-up time of all the instances in the dataset. The element $I_{ij}\ (i=1,\cdots,N; j=1,\cdots,K)$ of the indicator matrix will be $1$ if the event occurred before time $y_j$ for instance $i$, otherwise it will be $0$. One of the primary advantages of the MTLSA approach is that it can capture the dependency between the outcomes at various time points by using a shared representation across the related tasks in the transformation, which will reduce the prediction error on each task. In addition, the model can simultaneously learn from both uncensored and censored instances based on the indicator matrix. One important characteristic of non-recurring events, i.e., once the event occurs it will not occur again, is encoded via the \emph{non-negative non-increasing list} structure constraint. In the MTLSA algorithm, the $\ell_{2,1}$-norm penalty is employed to  learn a shared representation{, with time complexity $O(NPK)$,} across related tasks and  hence compute the relatedness between the individual models built for various unique event time points.

\section{Performance Evaluation Metrics}
\label{sec5}
Due to the presence of the censoring in survival data, the standard evaluation metrics for regression such as root of mean squared error and $R^2$ are not suitable for measuring the performance in survival analysis \cite{heagerty2005survival}. Instead, the prediction performance in survival analysis needs to be measured using more specialized evaluation metrics.
 
\subsection{C-index}
\begin{figure*}[!htp]
	\centering
	\scalebox{.6}{\includegraphics[trim={0cm 0.3cm 0cm 0.7cm},clip]{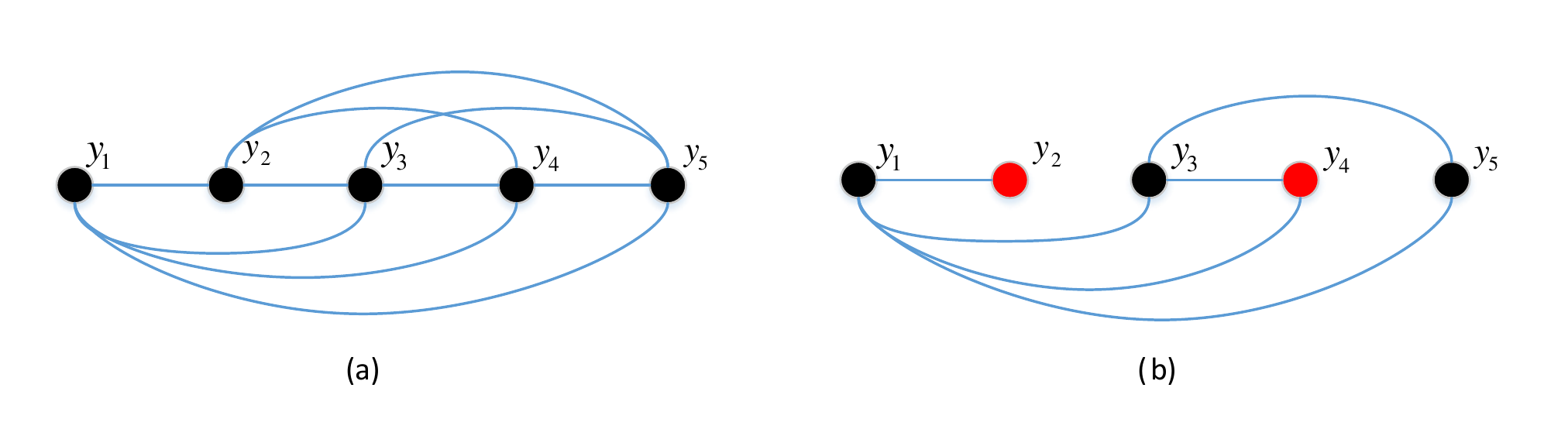}} \hspace{-0.1 cm}
	\caption{Illustration of the ranking constraints in survival data for C-index calculations ($y_1<y_2<y_3<y_4<y_5$). Here, black circles indicate the observed events and red circles indicate the censored observations. (a) No censored data and (b) With censored data.}
	\label{Fig:cindex}
	\vspace{-4mm} 
\end{figure*}

In survival analysis, a common way to evaluate a model is to consider the relative risk of an event for different instance instead of the absolute survival times for each instance. This can be done by computing the concordance probability or the concordance index (C-index)~\cite{harrell1984regression,harrell1982evaluating,pencina2004overall}. The survival times of two instances can be ordered for two scenarios: (1) both of them are uncensored; (2) the observed event time of the uncensored instance is smaller than the censoring time of the censored instance~\cite{steck2008ranking}. This can be visualized by the ordered graph given in Figure~\ref{Fig:cindex}. Figure~\ref{Fig:cindex}(a) and Figure~\ref{Fig:cindex}(b) are used to illustrate the possible ranking comparisons (denoted by edges between instances) for the survival data without and with censored instances, respectively. There are $\binom{5}{2}=10$ possible pairwise comparisons for the five instances in the survival data without censored cases shown in Figure~\ref{Fig:cindex}(a). Due to the presence of censored instances (represented by red circles) in Figure~\ref{Fig:cindex}(b), only $6$ out of the $10$ comparisons are feasible. It should be noted that, for a censored instance, only an earlier uncensored instance (for example $y_2\&y_1$) can be compared with. However, any censored instance cannot be compared with both censored and uncensored instances after its censored time (for example, $y_2\&y_3$ and $y_2\&y_4$) since its actual event time is unknown.

Consider both the observations and prediction values of two instances, $(y_1,\,\hat{y}_1)$ and $(y_2,\,\hat{y}_2)$, where $y_i$ and $\hat{y}_i$ represent the actual observation time and the predicted value, respectively. The concordance probability between them can be computed as 
\begin{equation}
c=Pr(\hat{y}_1>\hat{y}_2|y_1\geq y_2)
\end{equation}
By this definition, for the binary prediction problem, C-index will have a similar meaning to the regular area under the ROC curve (AUC), and if $y_i$ is binary, then the C-index is the AUC \cite{limulti}. As the definition above is not straightforward, in practice, there are multiple ways of calculating the C-index. 
\begin{enumerate}
\item When the output of the model is a hazard ratio (such as the outcome obtained by Cox based models), C-index can be computed using
\begin{equation}
\hat{c}=\frac{1}{num}\sum_{i:\delta_i=1}\sum_{j:y_i<y_j}I[X_i\hat{\beta}>X_j\hat{\beta}]
\end{equation}
where $i,j \in \{1,\cdots, N\}$, $num$ denotes the number of all comparable pairs, $I[\cdot]$ is the indicator function and $\hat{\beta}$ is the estimated parameters from the Cox based models.
\item For the survival methods which aim at directly learning the survival time, the C-index should be calculated as:
\begin{equation}
\hat{c}=\frac{1}{num}\sum_{i :\delta_i=1}\sum_{j:y_i<y_j}I[S(\hat{y}_j|X_j)>S(\hat{y}_i|X_i)]
\end{equation}
where $S(\cdot)$ is the estimated survival probabilities.
\end{enumerate}

%$S(\hat{H}_i|X_i)$ is the predicted hazard ratio (the patient who has a longer survival time should have a smaller hazard ratio).

%In 1982, Harrell et al. proposed the first definition of C-index for time-to-event data \cite{harrell1982evaluating}.  
%\begin{equation}
%\hat{c}_{harrell}:=\frac{\sum_{j<i}I(y_j<y_i)I(\hat{y}_j>\hat{y}_i)\Delta_j+I(y_i<y_j)I(\hat{y}_i>\hat{y}_j)\delta_i}{\sum_{j<i}I(y_j<y_i)\Delta_j+I(y_i<y_j)\delta_i}
%\end{equation}
%where $y_i, y_j$ and $\hat{y}_i,\hat{y}_j$ are the observed event times and the predicted values, respectively, for instances $i$ and $j$ with $i,j\in \{1,\cdots,N\}$ in the dataset. 
%However, this estimator of C-index ignores the case that the smaller observed survival time is censored, which lead it to a biased estimation. Uno et al. \cite{uno2011c}  proposed a modified version of $\hat{c}_{harrell}$ to overcome the censoring bias,
%\begin{equation}
%\hat{c}_{Uno}:=\frac{\sum_{i,j}(\hat{G}^L_n(y_i))^{-2}I(y_i<y_j)I(\hat{y}_i>\hat{y}_j)\delta_i}{\sum_{i,j}(\hat{G}^L_n(y_i))^{-2}I(y_i<y_j)\delta_i}
%\end{equation}
%where $\hat{G}^L_n(t)$ represents the Kaplan-Meier estimator of the survival function. and numerical studies shows the consistency and robust properties of this estimator. 
In order to evaluate the performance during a follow-up period, Heagerty and Zheng defined the C-index for a fixed follow-up time period $(0,t^*)$ as the weighted average of AUC values at all possible observation time points~\cite{heagerty2005survival}. %Based on the standard AUC definition given below
%\begin{equation}
%AUC=P(\hat{y}_i<\hat{y}_j|\delta_i=0,\delta_j=1)=\frac{1}{num}\sum_{\delta_i=0}\sum_{\delta_j=1}I(\hat{y}_i<\hat{y}_j)
%\end{equation}
The time-dependent AUC for any specific survival time $t$ can be calculated as
\begin{equation}
\label{tdauc}
AUC(t)=P(\hat{y}_i<\hat{y}_j|y_i<t,y_j>t)=\frac{1}{num(t)}\sum_{i:y_i<t}\sum_{j:y_j>t}I(\hat{y}_i<\hat{y}_j)
\end{equation}
where $t\in T_s$ which is the set of all possible survival times and $num(t)$ represents the number of comparable pairs for the time point $t$. Then the C-index during the time period $(0,t^*)$, which is the weighted average of the time-dependent AUC obtained by Eq.~(\ref{tdauc}), is computed as
%\begin{eqnarray}
%\nonumber
%c_{t^*}&=&\frac{1}{num}\sum_{i:\delta_i=1}\sum_{j:y_i<y_j}I(\hat{y}_i<\hat{y}_j)\\
%&=&\frac{1}{\sum\limits_{t \in T_s}num(t)}\sum_{t\in T_s}\sum_{y_i<t}\sum_{y_j>t}I(\hat{y}_i<\hat{y}_j)=\sum_{t\in T_s}AUC(t)\cdot\frac{num(t)}{num}
%\end{eqnarray}
\begin{eqnarray}
%\nonumber
c_{t^*}&=&\frac{1}{num}\sum_{i:\delta_i=1}\sum_{j:y_i<y_j}I(\hat{y}_i<\hat{y}_j)=\sum_{t\in T_s}AUC(t)\cdot\frac{num(t)}{num}
%&=&\frac{1}{num}\sum_{t\in T_s}num(t)[\frac{1}{num(t)}\sum_{i:y_i<t}\sum_{j:y_j>t}I(\hat{y}_i<\hat{y}_j)]=\sum_{t\in T_s}AUC(t)\cdot\frac{num(t)}{num}
\end{eqnarray}
Thus $c_{t^*}$ is the probability that the predictions are concordant with their outcomes for a given data during the time period $(0,t^*)$.
%In addition, in \cite{gonen2005concordance}, a asymptotically unbiased C-index which is specific for the Cox model was designed. Among these methods, Harrell et al.'s and Uno's c-index \cite{harrell1982evaluating} is suitable for all cases; in contrast, the C-index in \cite{heagerty2005survival} and \cite{gonen2005concordance} are designed specifically for the proportional hazards model, where $X_i\beta$ is used for computing of C-index instead of the estimated outcome $\hat{y}_i$.

\subsection{Brier Score}
Named after the inventor Glenn W. Brier, the Brier score (BS) \cite{brier1950verification} is developed to predict the inaccuracy of probabilistic weather forecasts. It can only evaluate the prediction models which have probabilistic outcomes; that is, the outcome must remain within the range [0,1], and the sum of all the possible outcomes for a certain individual should be 1. %For $Q$-class output problem (categorical outcome), the Brier score can be formulated as:
%\begin{equation}
%{BS}=\frac{1}{N}\sum_{i=1}^N\sum_{j=1}^Q[\hat{y}_{ij}-y_{ij}]^2
%\end{equation}
%where $y_{ij}$ and $\hat{y}_{ij}$ are the observed outcome and the predicted outcome with respect to class $j$ for instance $i$, respectively. They need to satisfy the constraints $\sum\limits_{j=1}^Q\hat{y}_{ij}=1$ and $\sum\limits_{j=1}^Qy_{ij}=1$. 
When we consider the binary outcome prediction with a sample of $N$ instances and for each $X_i\ (i=1,2,...,N)$, the predicted outcome at $t$ is $\hat{y}_i(t)$, and the actual outcome is $y_i(t)$; then, the empirical definition of the Brier score at the specific time $t$ can be given by %$BS(t)=\frac{1}{N}\sum_{i=1}^N[\hat{y}_i(t)-y_i(t)]^2$, 
\vspace{-0.9mm}
\begin{equation}
\vspace{-1.5mm}
\label{equation:brier0}
BS(t)=\frac{1}{N}\sum_{i=1}^N[\hat{y}_i(t)-y_i(t)]^2
%\vspace{-0.9mm}
\end{equation}
where the actual outcome $y_i(t)$ for each instance can only be $1$ or $0$. 

Brier score was extended in \cite{graf1999assessment} to be a performance measure for survival problems with censored information to evaluate the prediction models where the outcome to be predicted is either binary or categorical in nature. When incorporating the censoring information in the dataset, the individual contributions to the empirical Brier score are reweighted according to the censored information. Then, the Brier score can be updated as follows:
\vspace{-0.9mm} 
\begin{equation}
\label{equation:brier}
\vspace{-1.5mm}
{BS(t)}=\frac{1}{N}\sum_{i=1}^N w_i(t)[\hat{y}_i(t)-y_i(t)]^2
\end{equation}
In Eq.(\ref{equation:brier}), $w_i(t)$, given in Eq. (\ref{equation:weight}), denotes the weight for the $i^{th}$ instance and it is estimated by incorporating the Kaplan-Meier estimator of the censoring distribution $G$ obtained on the given dataset $(X_i,y_i,1-\delta_i),i=1,\cdots,N$.
\begin{equation}
	\label{equation:weight}
	%\nonumber
	w_i(t)=\begin{cases}
		\delta_i/G(y_i) &\text{if } y_i \le t \\
		1/G(y_i) &\text{if } y_i > t
	\end{cases}
\end{equation}
With this weight distribution, the weights for the instances that are censored before $t$ will be $0$. However, they contribute indirectly to the calculation of the Brier score since they are used for calculating $G$. The weights for the instances that are uncensored at $t$ are greater than $1$, so that they contribute their estimated survival probability to the calculation of the Brier score.

%When predictions are assessed over a time period $(0,t^*)$ instead of for a particular time point $t$, the prediction error can be averaged over the time interval by using the Integrated Brier Score (IBS)~\cite{graf1999assessment} as shown in Eq. (\ref{equation:intbrier}).%, in which the individual contributions to the Brier Score are reweighted according to the censoring information by incorporating the Kaplan-Meier estimator of the censoring distribution $G$ obtained based on $(y_i,1-\delta_i),i=1,\cdots,n$. 
%\begin{equation}
%\label{equation:intbrier}
%{IBS}=\frac{1}{N}\sum_{i=1}^N \int_{0}^{t^*}w_i(t)[\hat{y}_i(t)-y_i(t)]^2 dW(t)
%\end{equation}
%where $W(t)$ is a weight function, for which the natural choices are $W(t)=t/t^*$ or $W(t)=(1-\hat{S}(t))/(1-\hat{S}(t^*))$, where $\hat{S}(t)$ denotes the estimated survival function. From the definitions of Brier score, it is evident that it measures the mean squared difference between predictions made and the actual outcomes; therefore, the lower the Brier score, the better the prediction model.
\subsection{Mean Absolute Error}
For survival analysis problems, the mean absolute error (MAE) can be defined as an average of the differences between the predicted time values and the actual observation time values. It is calculated as follows:
\begin{equation}
MAE=\frac{1}{N}\sum_{i=1}^{N}(\delta_i|y_i-\hat{y}_i|)
\end{equation}
where $y_i$ $(i=1,\cdots,N)$ represents the actual observation times, and $\hat{y}_i$ $(i=1,\cdots,N)$ denotes the predicted times. It should be noted that only the samples for which the event occurs are being considered in this metric since if $\delta_i=0$, the corresponding term will become zero. MAE can only be used for the evaluation of survival models which can provide the event time as the predicted target value such as AFT models.
%\subsection{Hazard Ratio} The HR measure can consistently be used as a measure of misfit for predicted time of all survival models []. Before computing the Hazard Ratio (HR), all predicted time values should be normalized between 0 and 1. Then, the HR of true survival times are calculated under univariate Cox PH model (Introduced in section 2) using the normalized predicted time as input explanatory variables. 

%\subsection{Log-Rank $\chi^2$ Statistic} log-rank test is most commonly used for statistical comparison of survival times of two or more samples nonparametrically with data that are subject to censoring. In this measure, we take the median value of model outputs as a threshold, which is used to classify the test set into a low-risk or high-risk group. The log-rank test across these two groups is used to test how well the developed prognosis index splits between low-risk and high-risk patients. A large $\chi^2$-value for the test indicates that the model has been able to efficiently show risk differences between both groups.

\section{Related Topics} 
\label{sec6}
Besides the machine learning methods introduced in Section \ref{sec4} and the traditional statistical survival methods discussed in Section \ref{sec3}, there are few other topics that are closely related to survival analysis and we will summarize them now. 
 
\subsection{Early Prediction}
One of the primary challenges in the context of survival analysis, and in general longitudinal studies, is that a sufficient number of events in the training data can be collected only by waiting for a long period. This is the most significant difference from the regular supervised learning problems, in which the labels for each instance can be given by a domain expert in a reasonable time period. Therefore, a good survival model should have the ability to forecast the event occurrence at future time by using only a limited event occurrence information at the early stage of a {survival analysis} problem.

There are many real-world applications which motivate the need for new prediction models which can work using only the {data collected at the early stage of the studies}. For example, in the healthcare domain, it is critical to study the effect of a new treatment in order to understand the treatment or drug efficacy, which should be estimated as early as possible. In this case, the patients will be monitored over a certain time period and the event of interest will be the patient admission to the hospital due to the treatment failure. This scenario clearly indicates the need for algorithms which can predict the event occurrence effectively using only a few events. 

%Recently, most of other early prediction methods  \cite{xing2009early,ghalwash2013extraction,xing2011extracting} are based on time series classification methods. However, the early stage prediction problem discussed above is completely different from the time series forecasting problem since the goal here is to predict the outcome of (binary) event occurrence for each subject for a time which is much beyond the observation time (as opposed to merely predicting the next time step value which is typically done in the standard time series forecasting models). 

To solve this problem, an Early Stage Prediction (ESP) approach trained at early stages of {survival analysis} studies to predict the time-to-event is proposed in \cite{early2016}. Two algorithms based on Na\"{i}ve Bayes and Bayesian Networks 
%More specifically, three probabilistic algorithms based on Naive Bayes (NB), Tree-Augmented Naive Bayes (TAN) and Bayesian Network (BN), called ESP-NB, ESP-TAN and ESP-BN, respectively, 
are developed by estimating the posterior probability of event occurrence based on different extrapolation techniques using Weibull, Log-logistic and Log-normal distributions discussed in Section \ref{sec3}. The ESP framework is a two-stage algorithm: (1) Estimate the conditional probability distribution based on the training data collected until the early stage time point ($t_c$) of the study; (2) Extrapolating the prior probability of the event for the future time ($t_f$) using AFT model with different distributions. According to the experimental results in these works, the ESP framework can provide more accurate predictions when the prior probability at the future time is appropriately estimated using the current information of event occurrence. %This is extremely important in such longitudinal survival studies since accumulating enough training data about the event occurrence is a time-consuming process.

\subsection{Data Transformation}
In this section, we will discuss two data transformation techniques that will be useful for data pre-processing in survival analysis. Both of these approaches transform the data to a more conducive form so that other survival-based (or sometimes even the standard algorithms) can be applied effectively.
\subsubsection{Uncensoring approach}
In survival data, the 
%time to the event occurrence is not necessarily observed for all the instances and hence the outcome variable might be incomplete, which is called ‘censoring’. This 
incompleteness in the event (outcome) information makes it difficult for standard machine learning methods to learn from such data. The censored observations in survival data might look similar to unlabeled samples in classification or unknown response in regression problem in the sense that status or time-to-event is not known for some of the observations. However, different from unlabeled samples where the labeling information is completely missing, the censored instances actually have partial informative labeling information which provides the possible range of the corresponding true response (survival time). Such censored data have to be handled with special care within any machine learning method in order to make good predictions. Also, in survival analysis problems, only the information before a certain time point (before censoring occurs) is available for the censored instances and this information should be integrated into the prediction algorithm to obtain the most optimal result.

Typically, there are two naive ways of handling such censored data. One is to delete the censored instances, and it performs well if the number of the samples are large enough and the censoring instances are not censored randomly. However, it will provide a sub-optimal model because of neglecting the available information in those censored instances \cite{delen2005predicting,burke1997artificial}. Treating censoring as event-free is another naive and simple choice. This method performs well for data with only a few censored instances, but it underestimates the true performance of the model. Although these methods are simple for handling the censored data, they loose useful information available in the data. Here, we list two other approaches proposed in the literature to handle censored data. 
\begin{enumerate}
\item Group the instances in the given data into three categorizes \cite{zupan2000machine}: (i) instances which experience the event of interest during the observation will be labeled as event; (ii) instances whose censored time is later than a predefined time point are labeled as event-free; (iii) for instances whose censored time is earlier than a predefined time point, a copy of these instances will be labeled as event and another copy of the same instances will be labeled as event-free, respectively, and all these instances will be weighted by a marginal probability of event occurrence estimated by the Kaplan-Meier method. 

\item For each censored instance, estimate the probability of event and probability of being censored (considering censoring as a new event) using Kaplan-Meier estimator and give a new class label based on these probability values \cite{early2016}. For each instance in the data, when the probability of event exceeds the probability of being censored, then it is labeled as event; otherwise, it will be labeled as event-free which indicates that even if there is complete follow-up information for that instance, there is extremely low chance of event occurrence by the end of the observation time period.
%\item Learn separate models from observation time-divided data obtained by splitting the survival data into different groups and estimate the distribution of outcomes at different time for each group~\cite{biganzoli1998feed,jerez2003combined,zupan2000machine}. The distribution would be assessed through the outcome probability estimate based on the Kaplan-Meier method.

%\item Treating censoring in the survival data as class noise and performing a pre-processing denoising procedure to correct the outcomes of the more obvious censored positives. 
%@article{vstajduhar2012uncensoring,
%title={Uncensoring censored data for machine learning: A likelihood-based approach},
%author={{\v{S}}tajduhar, Ivan and Dalbelo-Ba{\v{s}}i{\'c}, Bojana},
%journal={Expert Systems with Applications},
%volume={39},
%number={8},
%pages={7226--7234},
%year={2012},
%publisher={Elsevier}
%}
\end{enumerate}

\subsubsection{Calibration}
%Right censoring is a common phenomenon that arises in many longitudinal studies where an event of interest could not be recorded within the given time frame.
Censoring causes missing time-to-event labels, and this effect is compounded when dealing with datasets which have high amounts of censored instances. Instead of using the uncensoring approach, calibration methods for survival analysis can also be used to solve this problem by learning more optimal time-to-event labels for the censored instances.
Generally, there are mainly two reasons which motivate calibration. First, the survival analysis model is built using the given dataset where the missing time-to-events for the censored instances are assigned to a value such as the duration of the study or last known follow up time. However, this approach is not suitable for handling data with many censored instances. In other words, for such data, these inappropriately labeled censored instances cannot provide much information to the survival algorithm. Calibration method can be used to overcome this missing time-to-events problem in survival analysis. Secondly, dependent censoring in the data, where censoring is dependent on the covariates, may lead to some bias in standard survival estimators, such as KM method. This motivates an imputed censoring approach which calibrates the time-to-event attribute to decrease the bias of the survival estimators.

In \cite{bhanu2016}, a calibration survival analysis method which uses a regularized inverse covariance based imputation is proposed to overcome the problems mentioned above. It has the ability to capture correlations between censored instances and correlations between similar features. In calibrated survival analysis, through imputing an appropriate label value for each censored instance, a new representation of the original survival data can be learned effectively. This approach fills the gap in the current literature by estimating the calibrated time-to-event values for these censored instances by exploiting row-wise and column-wise correlations among censored instances in order to effectively impute them.

\subsection{Complex Events}
Until now, the discussion in this paper has been primarily focused on survival problems in which each instance can experience only a single event of interest. However, in many real-world domains, each instance may experience different types of events and each event may occur more than once during the observation time period. For example, in the healthcare domain, one patient may be hospitalized multiple times due to different medical conditions. Since this scenario is more complex than the survival problems we discussed before, we consider them to be complex events. In this section, we will discuss two techniques, namely, competing risks and recurrent events, to tackle such complex events.   
%\vspace{-3mm}
\subsubsection{Competing Risks}
In the survival problem, if several different types of events are considered, but only one of them can occur for each instance over the follow-up period, then the competing risks will be defined as the probabilities of different events. In other words, the competing risks will only exist in survival problems with more than one possible event of interest, but only one event will occur at any given time. For example, in healthcare domain, a patient may have both heart attack and lung cancer before his death, but the reason of his death can be either lung cancer or heart attack, but not both. In this case, competing risks are the events that prevent an event of interest from occurring which is different from censoring. It should be noted that in the case of censoring, the event of interest still occurs at a later time, while the event of interest is impeded. %The competing risk model can be graphically represented with an initial state (alive or more generally, event-free) and a number of different events as the endpoints as shown in Figure \ref{Fig:competingrisk}. 
%\begin{figure*}[!htp]
%	\centering
%	%\scalebox{1}{\includegraphics[trim={0cm 0cm 0cm 0.5cm},clip]{figure/competingrisk}} \hspace{-0.1 cm}
%	\includegraphics[trim={0cm 0.3cm 0cm 0.3cm},clip]{figure/competingrisk.pdf}
%	\caption{Illustration of competing risks with $Q$ causes of failure.}
%	\label{Fig:competingrisk}
%\end{figure*}

%For the competing risk problem, the KM estimator may not be as informative as with only one event of interest since the independence assumption about the competing risks cannot be verified~\cite{kleinbaum2006survival}.
To solve this problem, the standard way is to analyze each of these events separately using the survival analysis approach by considering other competing events as censored~\cite{kleinbaum2006survival}. However, there are two primary drawbacks with such an approach. One problem is that this method assumes that the competing risks are independent of each other. In addition, it would be difficult to interpret the survival probability estimated for each event separately by performing the survival analysis for each event of interest in the competing risks. %Regarding the independence assumption, the typical competing risks analysis methods assume that this assumption holds even if this is not the case since it is hard to verify.

To overcome these drawbacks, two methods are developed in the survival analysis literature: Cumulative Incidence Curve (CIC) Approach and Lunn-McNeil (LM) Approach.
%\begin{enumerate}

\vspace{2mm}
\textbf{Cumulative Incidence Curve (CIC) Approach: }
To avoid the questionable interpretation problem, the cumulative incidence curve~\cite{putter2007tutorial} is one of the main approaches for competing risks which estimates the marginal probability of each event $q$. The CIC is defined as 
%\begin{equation}
%CIC(t_f)=\sum_{i=1}^f \hat{S}(t_{f-1})\hat{h}_c(t_f)
%\end{equation}
\begin{equation}
CIC_q(t)=\sum_{j:t_j\le t}\hat{S}(t_{j-1})\hat{h}_q(t_j)=\sum_{j:t_j\le t}\hat{S}(t_{j-1})\frac{n_{qj}}{n_j}
\end{equation}
where $\hat{h}_q(t_j)$ represents the estimated hazard at time $t_j$ for event $q\ (q=1,\cdots,Q)$, $n_{qj}$ is the number of events for the event $q$ at $t_j$, $n_j$ denotes the number of instances who are at the risk of experiencing events at $t_j$, and $S(t_{j-1})$ denotes the survival probability at last time point $t_{j-1}$. %An alternative to CIC is the conditional probability curve (CPC) \cite{kleinbaum2006survival}, which gives the probability of experiencing an event $c$ at time $t$, given that an individual has not experienced any of the other competing risks until time $t$. CPC can be computed from CIC as follows: $CPC_q={CIC_q}/{(1-CIC_{q^{\prime}})}$, where $CIC_{q^{\prime}}$ is the cumulative incidence of failure from risks other than risk $q$.

\vspace{2mm}
\textbf{Lunn-McNeil (LM) Approach }\cite{lunn1995applying}: It is an alternative approach to analyze the competing risks in the survival problems and it also allows the flexibility to conduct statistical inference from the features in the competing risk models. It fits a single Cox PH model which considers all the events in competing risks rather than separate models for each event~\cite{kleinbaum2006survival}. It should be noted that the LM approach is implemented using an augmented data, in which a dummy variable is created for each event to distinguish different competing risks. %The general form of LM approach is defined as:
%\begin{equation}
%h^*(t,{X})=h_{0}^*(t)\times exp[X_1\beta_{1}+\cdots+X_P\beta_P+\sum\limits_{j=2}^Q \sum_{i=1}^{P}D_jX_i\beta_{ji}]
%\end{equation}
%%\begin{eqnarray} 
%%h^*(t,{X})&=&h_{0}^*(t)\times exp[X_1\beta_{1}+\cdots+X_P\beta_P+\sum\limits_{j=2}^C \sum_{i=1}^{P}D_jX_i\beta_{ji}] \nonumber\\
%%&=&h_{0}^*(t)\times exp[X_1\beta_{1}+\cdots+X_P\beta_P+D_2X_1\beta_{21}+\cdots+D_2X_P\beta_{2P}   \nonumber\\
%%&+&D_3X_1\beta_{31}+\cdots+D_3X_P\beta_{3P}+\cdots+D_CX_1\beta_{C1}+\cdots+D_CX_P\beta_{CP}] 
%%\end{eqnarray}
%where $Q$ is the number of competing risks and $D_j$ is the dummy variable which equals to 1 for event $j$ and 0 otherwise for $\forall$ $j \in\{2,\cdots,Q\}$. We can see that there is no dummy variable for event $1$ since it is the reference event. For the competing risk $q\ (q=1,\cdots,Q)$, we have 
%\begin{equation}
%h_q^*(t,{X})=h_{0q}^*(t)exp[(\beta_1+\beta_{q1})X_1+\cdots+(\beta_1+\beta_{qP})X_P]
%\end{equation}
%Then the hazard ratio with respect to the effect of $X_1$ is given as follows: 
%\begin{equation}
%HR_i(X_1)=exp(\beta_1+\beta_{i1})	
%\end{equation}
%%LM method allows the flexibility to perform statistical inferences to determine whether a simpler version of an initial LM model is more appropriate.
%\end{enumerate}

\subsubsection{Recurrent Events}
In many application domains, the event of interest in survival problems may occur several times during the observation time period. This is significantly different from the death of the patients in healthcare domain. In such cases, the outcome event can occur for each instance more than once during the observation time period. In survival analysis, we refer to such events which occur more than once as \textit{recurrent events}, which contrasts with the competing risks discussed above. Typically, if all the recurring events for each instance are of the same type, the counting process (CP) algorithm \cite{andersen2012statistical} can be used to tackle this problem. If there are different types of events or the order of the events is the main goal, other methods using stratified Cox (SC) approaches can be used~\cite{ata2007cox}. These methods include stratified CP, Gap Time and Marginal approach. These approaches differ not only in the way they determine the risk set but also in the data format.

\vspace{2mm}
\textbf{Counting Process:} In Counting Process method, the data processing procedure is as follows: (i) For each instance, identify the time interval for each recurrent event and add one record to the data. It should be noted that an additional record for the event-free time interval should also be included for each instance. (ii) For each instance, each record of data should be labeled by a starting time and ending time of the corresponding time interval. 
%\begin{enumerate}
%\item Each instance contributes a record of data for each time interval corresponding to each recurrent event and any additional event-free follow-up interval.
%\item Each line of data for a given instance lists the start time and stop time for each interval of follow-up.
%\end{enumerate}
These properties of the data format distinguish the counting process method from other methods. They are significantly different from the regular survival data format for non-recurrent event problems, which provides only the ending time and contain only one record for each instance in the dataset. 

The key idea to analyze the survival data with recurrent events is to treat the different time intervals for each instance as independent records from different instances. The basic Cox model is used to perform the counting process approach. Each instance will not be removed from the risk set until the last time interval during the observation period. In other words, for the survival problem with recurrent events, the partial likelihood function formula is different from that in the non-recurrent event survival problems~\cite{kleinbaum2006survival}.

\vspace{2mm}
\textbf{Stratified Cox:} Stratified CP \cite{prentice1981regression}, Marginal \cite{wei1989regression} and Gap Time \cite{prentice1981regression} are three approaches using stratified Cox method to differentiate the event occurrence order. (1) In Stratified CP approach, the data format is exactly the same as that used in the CP approach, and the risk set for the future events is affected by the time of the first event. (2) In Marginal approach, it uses the same data format as the non-recurrent event survival data. This method considers the length of the survival time from the starting time of the follow-up until the time of a specific event occurrence and it assumes that each event is independent of other events. For the $k^{th}$ event $(k=1,2,\cdots)$ in this method, the risk set contains those instances which are at the risk of experience the corresponding event after their entry into the observation. (3) In the Gap Time approach, the data format (start, stop) is used, but the starting time for each data record is 0 and the ending time is the length of the interval from the previous experienced event. In this method, the risk set for the future events will not be affected by the time of the first event.
%\vspace{-2.2mm}
\section{Application Domains}
\label{sec7}
%It has become a common practice in many application domains to collect data over a period of time and record any events of interest that occur within this period. These studies are usually called longitudinal studies, in which the subjects are followed over a period of time for monitoring certain risks events. In such longitudinal studies, one of the main challenge is to estimate the occurrence of the event of interest. As an important analysis tool for such longitudinal data, survival analysis method can be applied to several real-world domains where the observations started from a particular starting time and will continue until the occurrence of a certain event or the observed objects become missing (not observed) from the study. 
In this section, we will demonstrate the applications of survival analysis in various real-world domains. Table \ref{tab:application} summarizes the events of interest, expected goal and the features that are typically used in each specific application described in this section. 
%\vspace{-1mm}
\begin{table}[h]
\setlength\extrarowheight{4pt}
\tbl{Summary of various real-world application domains where survival analysis was successfully used.\label{tab:application}}{%
\begin{tabular}{|c|c|c|c|}
\hline
\textbf{Application}& \textbf{Event of interest} &\textbf{Estimation} &\textbf{Features}\\\hline
\hline	
\thead{Healthcare\\\\ \cite{Kosiur01}\\ \cite{reddy2015review}} &\thead{Rehospitalization\\ Disease recurrence\\ Cancer survival} &\multicolumn{1}{m{2.5cm}|}{Likelihood of hospitalization within $t$ days of discharge.}  &\multicolumn{1}{m{5cm}|}{\textbf{Demographics}: age, gender, race. \textbf{Measurements}: height, weight, disease history, disease type, treatment, comorbidities, laboratory, procedures, medications.} \\\hline

\thead{Reliability\\\\ \cite{lyu1996handbook}\\ \cite{modarres2009reliability}} &{Device failure }  &\multicolumn{1}{m{2.5cm}|}{Likelihood of a device being failed within $t$ days.}  &\multicolumn{1}{m{5cm}|}{\textbf{Product}: model, years after production, product performance history. \textbf{Manufactory}: location, no. of products, average failure rate of all the products, annual sale of the product, total sale of the product. \textbf{User}: user reviews of the product.} \\\hline

\thead{Crowdfunding\\\\ \cite{rakesh2016probabilistic}\\ \cite{li2016project}  } &Project success & \multicolumn{1}{m{2.5cm}|}{Likelihood of a project being successful within $t$ days.} & \multicolumn{1}{m{5cm}|}{\textbf{Projects}: duration, goal amount, category. \textbf{Creators}: past success, location, no. of projects. \textbf {Twitter}: no. of promotions, backings, communities.  \textbf{Temporal}: no. of backers, funding, no. of retweets.}\\\hline

\thead{Bioinformatics\\\\ \cite{limulti}\\ \cite{beer2002gene}} & Cancer survival &\multicolumn{1}{m{2.5cm}|}{Likelihood of cancer within time $t$.}  & \multicolumn{1}{m{5cm}|}{\textbf{Clinical}: demographics, labs, procedures, medications. \textbf{Genomics}: gene expression measurements.}\\\hline

\thead{Student\\ Retention\\\\ \cite{murtaugh1999predicting}\\ \cite{ameri2016}} &Student dropout &\multicolumn{1}{m{2.5cm}|}{Likelihood of a student being dropout within $t$ days.} &\multicolumn{1}{m{5cm}|}{\textbf{Demographics}: age, gender, race. \textbf{Financial}: cash amount, income, scholarships.  \textbf{Pre-enrollment}: high-school GPA, ACT scores, graduation age. \textbf{Enrollment}: transfer credits, college, major. \textbf{Semester performance}: semester GPA, \% passed credits, \% dropped credits.} \\\hline

\thead{Customer \\Lifetime Value\\\\ \cite{zeithaml2001driving}\\ \cite{berger1998customer}} & Purchase behavior &\multicolumn{1}{m{2.5cm}|}{Likelihood of a customer purchasing from a given service supplier within $t$ days.} &\multicolumn{1}{m{5cm}|}{\textbf{Customer}: age, gender, occupation, income, education, interests, purchase history. \textbf{Store/Online store}: location, customer review, customer service, price, quality, shipping fees and time, discount.} \\\hline
	
\thead{Click \\Through Rate\\\\ \cite{yin2013silence}\\ \cite{barbieri2016improving}} & User clicking&\multicolumn{1}{m{2.5cm}|}{Likelihood of a user clicking the advertisement within time $t$.}  &\multicolumn{1}{m{5cm}|}{\textbf{User}: gender, age, occupation, interests, user’s click history. \textbf{Advertisement (ad)}: time of the ad, location of the ad on the website, topics of the ad, ad format, total click times of the ad. \textbf{Website}: no. of users of the website, page view each day of the website, no. of websites linking to the website.} \\\hline
		
\thead{Unemployment \\Duration in\\ Economics\\ \\ \cite{kiefer1988economic}} &Getting a job &\multicolumn{1}{m{2.5cm}|}{Likelihood of a person finding a new job within $t$ days.} &\multicolumn{1}{m{5cm}|}{\textbf{People}: age, gender, major, education, occupation, work experience, city, expected salary. \textbf{Economics}: job openings, unemployment rates every year.} \\\hline					

\end{tabular}}
%\vspace{-2mm}
\end{table}

\subsection{Healthcare}
In the healthcare domain, the starting point of the observation is usually a particular medical intervention such as a hospitalization admission, the beginning of taking a certain medication or a diagnosis of a given disease \cite{klein2005survival,Kosiur01}. The event of interest might be death, hospital readmission, discharge from the hospitalization or any other interesting incident that can happen during the observation period. The missing trace of the observation is also an important characteristic of the data collected in this domain. For example, during a given hospitalization, some patients may be moved to another hospital and in such cases, that patient will become unobserved from the study with respect to the first hospital after that time point. 
%Survival analysis is useful whenever the goal is to estimate the time for an event occurrence. 
In healthcare applications, survival prediction models primarily aim at estimating the failure time distribution and the prognostic evaluation of different features, including histological, biochemical and clinical characteristics \cite{marubini2004analysing}.
%\vspace{-1mm}

\subsection{Reliability}
In the field of reliability, it is a common practice to collect data over a period of time and record the interesting events that occur within this period. Reliability prediction focuses on developing methods which are good at accurately estimating the reliability of the new products~\cite{modarres2009reliability,lyu1996handbook}. The event of interest here corresponds to the time taken for a device to fail. In such applications, it is desirable to be able to estimate which devices will fail and if they do, when they will fail. Survival analysis methods can help in building such prediction models using the available information about these devices. These models can provide early warnings about potential failures, which is significantly important to either prevent or reduce the likelihood of failures and to identify and correct the causes of device failures.

\subsection{Crowdfunding}
%\cite{rakesh2015project,rakesh2016probabilistic,li2016project}
In recent years, the topic of crowdfunding has gained a lot of attention. Although the crowdfunding platforms have been successful, the percentage of the projects which achieved their desired goal amount is less than 50\% \cite{rakesh2015project}. Moreover, many of the prominent crowdfunding platforms follow the ``all-or-nothing" policy. In other words, if the goal is achieved before the pre-determined time period, the pledged funding can be collected. Therefore, in the crowdfunding domain, one of the most important challenges is to estimate the success probability of each project. %If the project success is correctly estimated, it will provide some guideline to both the project creators and backers about the progress and potential of the project.
%However, merely estimating whether a project will be successful or not using its corresponding goal date cannot provide a proper guideline to the backers who want to invest in popular projects. 
The need to estimate the project success probability motivates the development of new prediction approaches which can integrate the advantages of both regression (for estimating the time for success) and classification (for considering both successful and failed projects simultaneously in the model)~\cite{li2016project}. 
%There exist some difficulties to model the crowdfunding data for prediction purpose since the information about each project is changing dynamically and both the failed and successful projects are collected in the data. 
For the successful projects, the time to the success can be collected easily. However, for the projects that failed, it is not possible to collect the information about the length of the time for project success. The only information that can be collected is the funding amount that they raised until the pre-determined project end date. The authors in \cite{li2016project} consider both the failed and successful projects simultaneously by using censored regression methods. It fits the probability of project success with log-logistic and logistic  distributions and predicts the time taken for a project to become potentially successful. %This is significantly different from the standard regression method which only uses successful projects for training. 
%\vspace{-3.5mm}

\subsection{Bioinformatics}
One of the most popular applications of survival analysis in the domain of bioinformatics is gene expression. Gene expression is the process of synthesizing a functional gene product from the gene information and can be quantified by measuring either message RNA (mRNA) or proteins. Gene expression profiling is developed as a powerful technique to study the cell transcriptome. In recent years, multiple studies~\cite{limulti,beer2002gene} have correlated gene expression with survival outcomes in cancer applications in a genome-wide scale. Survival analysis methods are helpful in assessing the effect of single gene on survival prognosis and then identifying the most relevant genes as biomarkers for patients. In this scenario, the event of interest is the specific type of cancer (or any disease), and the goal is to estimate the likelihood of cancer using the gene expression measurements values. Generally, the survival prediction based on gene expression data is a high-dimensional problem since each cell contains tens of thousands of mRNA molecules. The authors in~\cite{antonov2014ppisurv} developed a statistical tool for biomedical researchers to define the clinical relevance of genes under investigation via their effect on the patient survival outcome. Survival analysis methods have shown to be effective in predicting the gene expression for different cancer data with the censored information.

\subsection{Student Retention}
In higher education, student retention rate can be evaluated by the percentage of students who return to the same university for the following semester after completing a semester of study. In the U.S. and around the world, one of the long-term goals of a university is to improve the student retention. Higher the student retention rate, more probable for the university to be positioned higher, secure more government funds, and have an easier path to program accreditations. In view of these reasons, directors and administrators in higher education constantly try to implement new strategies to increase student retention. Survival analysis has success in student retention problem~\cite{murtaugh1999predicting,ameri2016}. The goal of survival analysis is to estimate the time of event occurrence, which is critical in student retention problems because both correctly identifying whether a student will dropout and estimating when the dropout will happen are important. In such cases, it will be helpful if one can reliably estimate the dropout risk at the early stage of student education using both pre-enrollment and post-enrollment information. %We can take advantage of the survival analysis method at early stage of college study to predict student success.
%\vspace{-3.5mm}
%\subsection{User Behavior Modeling}
\subsection{Customer Lifetime Value}
Customer lifetime value (LTV) \cite{berger1998customer,zeithaml2001driving} of a customer refers to the profit that the customer brings to the store based on the purchase history. In the marketing domain, the customer LTV is used to evaluate the relationships between the customers and the store. It is important for a store to improve the LTV in order to maintain or increase its profits in the long term since it is often quite expensive to acquire new customers. In this case, the main goal of this problem is to identify purchase patterns of the  customers who have a high LTV and provide recommendations for a relatively new user who has similar interest. Identifying loyal customers using LTV estimation has been studied by various researchers~\cite{rosset2003customer,mani1999statistics} using survival analysis methods and data mining approaches which are helpful in identifying the purchase patterns. Then, LTV will be defined using a survival function, which can be used to estimate the time of purchase for every customer from a given store, using the store information and also the available customer demographic information in the store database, such as the gender, income and age. 
%\vspace{-5mm}
\subsection{Click-Through Rate}
Nowadays, many free web services, including online news portals, search engines and social networks present users with advertisements \cite{barbieri2016improving}. Both the topics and the display orders of the ads will affect the user clicking probability~\cite{richardson2007predicting}.
%Generally, a matching system retrieves all the ads which are similar to the content of users' query first, and then all eligible ads will be ordered using a score-based function which combines the amount of money that the advertiser is willing to pay for its ad to be shown and the quality of the matching. This mechanism helps web services to maintain the quality of the ads they provide to the users, including being able to control the experience of users when they click on an ad. 
Studies have mostly focused on predicting the click-through rate (CTR) which indicates the percentage of the users who click on a given ad. It can be calculated as the ratio of the clicking times and the corresponding presentation times (no. of ad impressions). The CTR value indicates the attraction effect of the ad to the users~\cite{barbieri2016improving}. 
%There are mainly two categories of research works with regards to CTR, namely, pre-click and post-click. In the pre-click problem, 
The goal is to predict how likely the user will click the ads based on the available information about the website, users and ads. The time taken to click the ad is considered to be the event time. Those users who did not click on the ads are considered to be censored observations. %The authors in \cite{barbieri2016improving} estimated the post-click engagement on native ads by predicting the dwell time on the corresponding ad landing pages using survival analysis methods. They used dwell time \cite{yin2013silence}, defined as the actual length of time that a user spends on the ad landing page before leaving it, as a measure of engagement for the ads. In this problem, the event of interest corresponds to the return of the user to the main page, after he or she visited the landing page associated with an ad, and the dwell time is considered to be the survival time.
%\vspace{-3.5mm}
\subsection{Duration Modeling in Economics}
Traditionally, duration data which measures how long individuals remain in a certain state is analyzed in biometrics and medical statistics using survival analysis methods. Actually, duration data also appears in a wide variety of situations in economics such as unemployment duration, marital instability and time-to-transaction in the stock market. Among them, the unemployment duration problem, which is the most widely studied one, analyzes the time people spend without a job~\cite{gamerman1987}. Generally, in the domain of economics, the time of being unemployed is extremely important since the length of unemployment of people plays a critical role in economics theories of job search \cite{kiefer1988economic}. For this problem, the data contains information on the time duration of unemployment for each individual in the sample. The event of interest here is getting a new job for each person and the objective is to predict the likelihood of getting a new job within a specific time period. It is desirable to understand how the re-employment probability changes over the period of the spell and to know more about the effect of the unemployment benefits on these probabilities.

%\vspace{-1mm}
\section{Resources}
\label{sec8}
This section provides a list of software implementations developed in various statistical methods and machine-learning algorithms for survival analysis. Table \ref{tab:summary} summarizes the basic information of the software packages for each survival method. We can find that most of the existing survival analysis methods can be implemented in R. %Now we will briefly introduce the usage of these packages.
\begin{table}[!ht]
	\centering
	\setlength\extrarowheight{3pt}
	\tbl{Summary of software packages for various survival analysis methods.\label{tab:summary}}{%
		\begin{tabular}{|c|c|c|l|}%{|p{2cm}|p{2cm}|p{2cm}|p{8cm}|}
			\hline
			\textbf{Algorithm}  &\textbf{Software} & \textbf{Language} &\textbf{Link}\\\hline
			\hline
			%non-parametric	
			Kaplan-Meier&\multirow{3}[0]{*}{survival} & \multirow{3}[0]{*}{R} & \multirow{3}[0]{*}{https://cran.r-project.org/web/packages/survival/index.html}\\\cline{1-1}
			Nelson-Aalen & &  & \multicolumn{1}{m{3cm}|}{}\\\cline{1-1}
			Life-Table & &  & \multicolumn{1}{m{3cm}|}{}\\
			\specialrule{1.5pt}{1pt}{1pt}
			
			%semi-parametric
			Basic Cox&\multirow{2}[0]{*}{survival} &\multirow{2}[0]{*}{R} &\multirow{2}[0]{*}{https://cran.r-project.org/web/packages/survival/index.html} \\\cline{1-1}
			TD-Cox& & & \\\hline
			Lasso-Cox& \multirow{3}[0]{*}{fastcox}& \multirow{3}[0]{*}{R}&\multirow{3}[0]{*}{https://cran.r-project.org/web/packages/fastcox/index.html} \\\cline{1-1}
			Ridge-Cox& & & \\\cline{1-1}
			EN-Cox& & & \\\hline
			Oscar-Cox&{RegCox} &{R}&{https://github.com/MLSurvival/RegCox}\\\hline
			CoxBoost& CoxBoost&R & \multicolumn{1}{m{3cm}|}{https://cran.r-project.org/web/packages/CoxBoost/}\\
			\specialrule{1.5pt}{1pt}{1pt}
			
			%parametric
			Tobit &survival& R &https://cran.r-project.org/web/packages/survival/index.html \\\hline
			BJ &bujar & R& \multicolumn{1}{m{3cm}|}{https://cran.r-project.org/web/packages/bujar/index.html}\\\hline
			%\thead{Weighted \\Regression} & & & \multicolumn{1}{m{3cm}|}{}\\\hline
			%\thead{Structured \\Regularization} & & & \multicolumn{1}{m{3cm}|}{}\\\hline
			AFT& survival& R &https://cran.r-project.org/web/packages/survival/index.html \\
			\specialrule{1.5pt}{1pt}{1pt}
			
			%machine learning
			\thead{Baysian \\Methods}& BMA &R & \multicolumn{1}{m{3cm}|}{https://cran.r-project.org/web/packages/BMA/index.html}\\\hline
			RSF&\scriptsize{randomForestSRC} & R& \multicolumn{1}{m{3cm}|}{https://cran.r-project.org/web/packages/randomForestSRC/}\\\hline
			BST& ipred &R & \multicolumn{1}{m{3cm}|}{https://cran.r-project.org/web/packages/ipred/index.html}\\\hline
			Boosting&mboost & R& \multicolumn{1}{m{3cm}|}{https://cran.r-project.org/web/packages/mboost/}\\\hline
			\thead{Active\\ Learning}&RegCox & R& \multicolumn{1}{m{3cm}|}{https://github.com/MLSurvival/RegCox}\\\hline
			\thead{Transfer\\ Learning}&TransferCox &C++ & \multicolumn{1}{m{3cm}|}{https://github.com/MLSurvival/TransferCox}\\\hline
			\thead{Multi-Task\\ Learning}& MTLSA & Matlab& https://github.com/MLSurvival/MTLSA\\
			%BoostCI&--- &R & {files.figshare.com/1339232/Text S1.pdf}\\\hline
			\specialrule{1.5pt}{1pt}{1pt}
			
			%related topics
			Early Prediction&\multirow{2}[0]{*}{ESP} &\multirow{2}[0]{*}{R} & \multirow{2}[0]{*}{https://github.com/MLSurvival/ESP}\\\cline{1-1}
			Uncensoring&{} &{} & {}\\\hline
			Calibration&survutils &R &https://github.com/MLSurvival/survutils \\\hline
			
			\thead{Competing Risks}&\multirow{1}[0]{*}{survival} & \multirow{1}[0]{*}{R} &\multirow{1}[0]{*}{https://cran.r-project.org/web/packages/survival/index.html}\\\hline
			%Competing Risks& & & \\\hline
			Recurrent Events&survrec &R &https://cran.r-project.org/web/packages/survrec/ \\\hline
			%Nerual Network& & & \multicolumn{1}{m{3cm}|}{}\\\hline
			%SVM& & & \multicolumn{1}{m{3cm}|}{}\\\hline				
		\end{tabular}}
		\vspace{-1mm}
\end{table}%

\begin{enumerate}
\item \emph{Non-parametric methods}: All the three non-parametric survival analysis methods can be implemented by employing the function \emph{coxph} and \emph{survfit} in the \emph{survival} package in R.

\item \emph{Semi-parametric methods}: Both the basic Cox model and the time-dependent Cox model can be trained using \emph{coxph} function in the \emph{survival} package in R. The Lasso-Cox, Ridge-Cox and EN-Cox in the regularized Cox methods can be trained using the \emph{cocktail} function in the \emph{fastcox} package. The \emph{RegCox} package can be used to implement OSCAR-Cox method. The \emph{CoxBoost} function in the \emph{CoxBoost} package can fit a Cox model by likelihood based boosting algorithm.

\item \emph{Parametric methods}: The Tobit regression can be trained using the \emph{survreg} function in the \emph{survival} package. Buckley-James Regression can be fitted using the \emph{bujar} package. The parametric AFT models can be trained using the \emph{survreg} function with various distributions.

\item \emph{Machine learning methods}: The \emph{BMA} package can be used to train a Bayesian model by averaging for Cox models. Bagging survival tree methods can be implemented using the \emph{bagging} function in the R package \emph{ipred}. Random survival forest is implemented in the \emph{rfsrc} function in the package \emph{randomForestSRC}. The \emph{mboost} function in the package \emph{mboost} can be used to implement the boosting algorithm. The \emph{arc} function in \emph{RegCox} package can be used to train the active learning survival model. Transfer-Cox model is written in C++ language. The multi-task leaning survival method is implemented using \emph{MTLSA} package in MATLAB.

\item \emph{Related topics}: The \emph{ESP} package which performs the early stage prediction for survival analysis problem also incorporates the uncensoring functions in the data pre-processing part. The \emph{survutils} package in R can be used to implement the calibration for the survival datasets.
%The \emph{survival} package can also be used to perform the residual analysis using two functions \emph{residuals.coxph}  and \emph{residuals.survreg}.
The \emph{survfit} function in \emph{survival} package can also be used to train the model for competing risks. In addition, the function \emph{Survr} in package \emph{survrec} can be used to train the survival analysis model for recurrent event data.
\end{enumerate}

\section{Conclusion}
\label{sec9}
The primary goal of survival analysis is to predict the occurrence of specific events of interest at future time points. Due to the widespread availability of {survival} data from various real-world domains combined with the recent developments in various machine learning methods, there is an increasing demand for understanding and improving methods for effectively handling survival data. In this survey article, we provided a comprehensive review of the conventional survival analysis methods and various machine learning methods for survival analysis, and described other related topics along with the evaluation metrics. We first introduced the basic notations and concepts in survival analysis, including the structure of survival data and the common functions used in survival analysis. Then, we introduced the well-studied statistical survival methods and the representative machine learning based survival methods. Furthermore, the related topics in survival analysis, including data transformation, early prediction and complex events, were also discussed. We also provided the implementation details of these survival methods and described the commonly used performance evaluation metrics for these models. Besides the traditional applications in healthcare and biomedicine, survival analysis was also successfully applied in various real-world problems, such as reliability, student retention and user behavior modeling.  
%\vspace{-2mm}

% Appendix
%\appendix
%\section*{APPENDIX}
%\setcounter{section}{1}
%In this appendix, we measure the channel switching time of Micaz [CROSSBOW] sensor devices. In our experiments, one mote alternatingly switches between Channels 11 and 12. Every time after the node switches to a channel, it sends out a packet immediately and then changes to a new channel as soon as the transmission is finished. We measure the number of packets the test mote can send in 10 seconds, denoted as $N_{1}$. In contrast, we also measure the same value of the test mote without switching channels, denoted as $N_{2}$. We calculate the channel-switching time $s$ as
%\begin{eqnarray}%
%s=\frac{10}{N_{1}}-\frac{10}{N_{2}}. \nonumber
%\end{eqnarray}%
%By repeating the experiments 100 times, we get the average channel-switching time of Micaz motes: 24.3$\mu$s.

%\appendixhead{ZHOU}

% Acknowledgments
%\begin{acks}
%This material is based upon work supported by, or in part by, the U.S. National Science Foundation grants IIS-1231742, IIS-1527827 and IIS-1646881.
%\end{acks}

% Bibliography
\bibliographystyle{ACM-Reference-Format-Journals}
\bibliography{survey2017}
                             % Sample .bib file with references that match those in
                             % the 'Specifications Document (V1.5)' as well containing
                             % 'legacy' bibs and bibs with 'alternate codings'.
                             % Gerry Murray - March 2012

% History dates
%\received{February 2007}{March 2009}{June 2009}

% Electronic Appendix
%\elecappendix

\medskip

%\section*{Appendix}
%\renewcommand\thesection{\Alph{section}}
%\section{Resources}
%The aim of this Appendix is to provide a useful list of software that implements survival methods. Table () summarizes the features of the main software packages.
%\begin{table}
%\tbl{Software for Survival Analysis Models\label{tab:software}}{%
%\begin{tabular}{|c|c|c|c|}
%\hline
%Method & Software &Language &S\\\hline	
%\hline
%Kaplan-Meier& &  &\\\hline
%Life-Table& &  &\\\hline
%\end{tabular}}
%\end{table}%

\end{document}